\documentclass{article}

\PassOptionsToPackage{numbers, compress, sort}{natbib}

\usepackage[final]{neurips_2022}

\usepackage[utf8]{inputenc} 
\usepackage[T1]{fontenc}    
\usepackage{hyperref}       
\usepackage{url}            
\usepackage{booktabs}       
\usepackage{amsfonts}       
\usepackage{amsmath}
\usepackage{amssymb}   
\usepackage{nicefrac}       
\usepackage{microtype}      
\usepackage{xcolor}         
\usepackage{colortbl}
\usepackage{multicol}
\usepackage{multirow}
\usepackage{graphicx}
\usepackage{wrapfig}
\usepackage{caption}
\usepackage{float}
\usepackage[ruled,vlined]{algorithm2e}

\usepackage{array, blkarray, makecell}
\definecolor{aliceblue}{rgb}{0.94, 0.97, 1.0}
\definecolor{applegreen}{rgb}{0.55, 0.71, 0.0}
\definecolor{debianred}{rgb}{0.84, 0.04, 0.33}
\definecolor{mintgreen}{rgb}{0.6, 1.0, 0.6}
\definecolor{honeydew}{rgb}{0.94, 1.0, 0.94}
\definecolor{malachite}{rgb}{0.04, 0.85, 0.32}
\definecolor{lime}{rgb}{0.75, 1.0, 0.0}

\newcommand\perfinc[1]{\textcolor{applegreen}{\small{~(+#1)}}}
\newcommand\perfdec[1]{\textcolor{debianred}{\small{~(-#1)}}}

\definecolor{commentcolor}{RGB}{110,154,155}   
\newcommand{\PyComment}[1]{\ttfamily\textcolor{commentcolor}{\# #1}}  
\newcommand{\PyCode}[1]{\ttfamily\textcolor{black}{#1}} 

\newcommand{\modelname}{3DTRL}

\title{Learning Viewpoint-Agnostic Visual Representations by Recovering Tokens in 3D Space}

\author{%
  Jinghuan~Shang$^1$~\quad~Srijan~Das$^{1,2}$~\quad~Michael~S.~Ryoo$^1$\\
  Department of Computer Science\\
  $^1$Stony Brook University, $^2$University of North Carolina at Charlotte\\
  $^1$\texttt{\{jishang, mryoo\}@cs.stonybrook.edu, $^2$sdas24@uncc.edu}\\
}
\begin{document}
\maketitle
\begin{abstract}
 Humans are remarkably flexible in understanding viewpoint changes due to visual cortex supporting the perception of 3D structure. In contrast, most of the computer vision models that learn visual representation from a pool of 2D images often fail to generalize over novel camera viewpoints. Recently, the vision architectures have shifted towards convolution-free architectures, visual Transformers, which operate on tokens derived from image patches. However, these Transformers do not perform explicit operations to learn viewpoint-agnostic representation for visual understanding. To this end, we propose a 3D Token Representation Layer (\modelname) that estimates the 3D positional information of the visual tokens and leverages it for learning viewpoint-agnostic representations. 
 The key elements of \modelname~include a pseudo-depth estimator and a learned camera matrix to impose geometric transformations on the tokens, trained in an unsupervised fashion. These enable \modelname~to recover the 3D positional information of the tokens from 2D patches. 
 In practice, \modelname~is easily plugged-in into a Transformer. Our experiments demonstrate the effectiveness of \modelname~in many vision tasks including image classification, multi-view video alignment, and action recognition. The models with \modelname~outperform their backbone Transformers in all the tasks with minimal added computation. Our code is available at \url{https://github.com/elicassion/3DTRL}.
\end{abstract}

\section{Introduction}
Over the past few years, computer vision models have developed rapidly from CNNs~\cite{vgg16,resnet,inception} to now Transformers~\cite{dosovitskiy2020vit,deit,tnt}. With these models, we can now accurately classify objects in an image, align image frames among video pairs, classify actions in videos, and more. Despite their successes, many of the models neglect that the world is in 3D and do not extend beyond the XY image plane~\cite{meshrcnn}. While humans can readily estimate the 3D structure of a scene from 2D pixels of an image, most of the existing vision models with 2D images do not take the 3D structure of the world into consideration. This is one of the reasons why humans are able to recognize objects in images and actions in videos regardless of their viewpoint, but the vision models often fail to generalize over novel viewpoints~\cite{NPL_2021_CVPR,viewclr,meshrcnn}.

Consequently, in this paper, we develop an approach to learn viewpoint-agnostic representations for a robust understanding of the visual data. 
Naive solutions to obtain viewpoint-agnostic representation would be either supervising the model with densely annotated 3D data, or learning representation from a large scale 2D datasets with samples encompassing different viewpoints.
Given the fact that such high quality data are expensive to acquire and hard to scale, an approach with a higher sample efficiency without 3D supervision is desired.

To this end, we propose a 3D Token Representation Layer (\modelname), incorporating 3D camera transformations into the recent successful visual Transformers~\cite{dosovitskiy2020vit,deit,crossvit,liu2021Swin}. 
\modelname~first recovers camera-centered 3D coordinates of each token by depth estimation.
Then \modelname~estimates a camera matrix to transform these camera-centered coordinates to a 3D world space.
In this world space, 3D locations of the tokens are absolute and view-invariant, which contain important information for learning viewpoint-agnostic representations.
Therefore, \modelname~incorporates such 3D positional information in the Transformer backbone in the form of 3D positional embeddings, and generates output tokens with the 3D information. 
Unlike visual Transformers only relying on 2D positional embedding, models with \modelname~are more compliant with learning viewpoint-agnostic representations.

We conduct extensive experiments on various vision tasks to confirm the effectiveness of \modelname. Our \modelname~outperforms the Transformer backbones on 3 image classification, 5 multi-view video alignment, and 2 multi-view action recognition datasets in their respective tasks. Moreover, \modelname~is a light-weighted, plug-and-play module that achieves the above improvements with minimal (2\% computation and 4\% parameters) overhead.

In summary, we present a learnable, differentiable layer \modelname~that efficiently and effectively learns viewpoint-agnostic representations.
\section{Background: Pinhole Camera Model} \label{background}
\modelname~is based on the standard pinhole camera model widely used in computer vision. Thus, we first briefly review the pinhole camera model. In homogeneous coordinate system, given a point with world coordinate $p^{\text{world}}$, a camera projects a pixel at $p$ on an image by:
\begin{align}
    p = \mathbf{K}~[~\mathbf{R}|\mathbf{t}~]~p^{\text{world}},
\end{align}
where $\mathbf{K}$ is the intrinsic matrix and $[~\mathbf{R}|\mathbf{t}~]$ is the extrinsic matrix. $\mathbf{K}$ is further represented by
\begin{align}
    \mathbf{K} = \begin{bmatrix}
        c & 0 & u_0 \\
        0 & c & v_0 \\
        0 & 0 & 1
    \end{bmatrix},
\end{align}
where $c$ is the focal length and $(u_0, v_0)$ are offset.
In this work, we explore visual understanding in multi-view setting, thus aiming at learning viewpoint-agnostic representations. In this setting, a scene may be captured with different cameras positioned at non-identical locations and viewing angles. 
Here, the world coordinate $p^{\text{world}}$ is the same across all the cameras while pixel projection $p$ is different across cameras. 
We focus on how to estimate the world coordinates $p^{\text{world}}$ from their corresponding image pixels at $p$.
Estimating $p^{\text{world}}$ from $p$ involves two transformations that correspond to the \textit{inverse} of $\mathbf{K}$ and $[~\mathbf{R}|\mathbf{t}~]$ which might not be known beforehand. Thus, we learn to \textit{estimate} them from image patches instead, which is a key procedure in \modelname.

\section{3D Token Representation Layer (\modelname)}
\begin{figure}
    \centering
    \includegraphics[width=\textwidth]{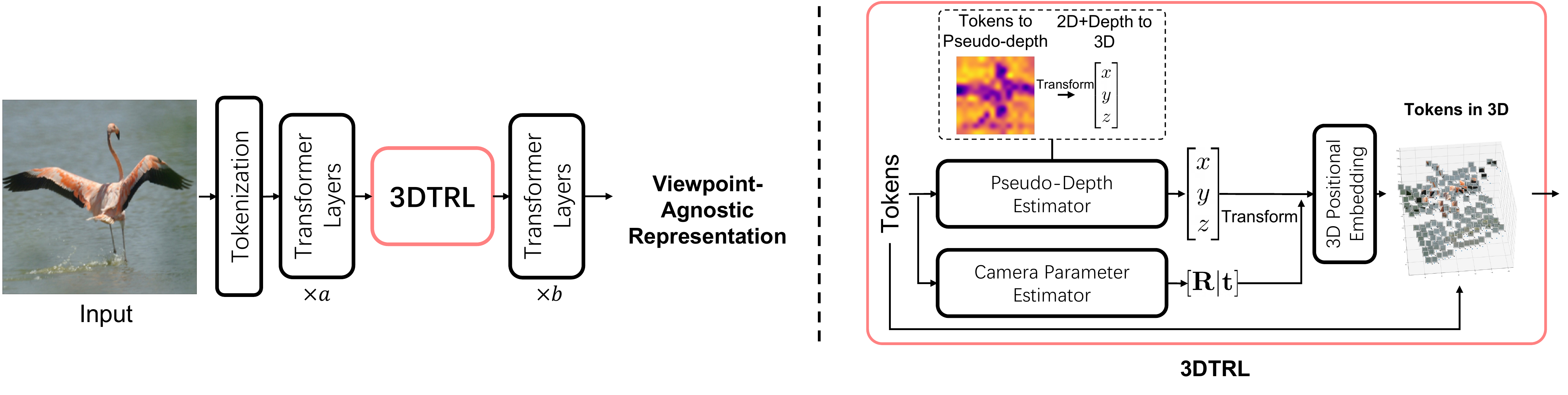}
    \caption{Overview of proposed \modelname. \textbf{Left}: \modelname~is a module inserted in between Transformer layers. 
    \textbf{Right}: \modelname~has three parts, Pseudo-Depth Estimator, Camera Parameter Estimator, and 3D Positional Embedding layer. Within Pseudo-Depth Estimator, we first estimate depth of each token and then calculate 3D coordinates from 2D locations and depth.}
    \label{fig:overview}
\end{figure}
In this section, we detail how \modelname~estimates 3D positional information in a Transformer and is integrated. We will introduce \modelname~for image analysis, then we adapt it to video models. 

\subsection{Overview}
\modelname~is a simple yet effective plug-in module that can be inserted in between the layers of visual Transformers (Figure~\ref{fig:overview} left). Given a set of input tokens $S$, \modelname~returns a set of tokens with 3D information. The number of tokens and their dimensionality are kept unchanged. 
Within the module (Figure~\ref{fig:overview} right), \modelname~first performs 3D estimation using two estimators: (1) pseudo-depth estimator and (2) camera parameter estimator. Then the tokens are associated with their recovered world 3D coordinates, which are transformed from estimated depth and camera matrix.
Finally, \modelname~generate 3D positional embeddings from these world coordinates and combine them with input $S$ to generate the output of \modelname. 

In the aspect that we insert \modelname~in between the Transformer model, \modelname~implicitly leverages the Transformer layers \textit{before} it to be a part of the 3D estimators, and layers \textit{after} that to be the actual 3D feature encoder (Figure~\ref{fig:overview} left). This avoids adding a large number of parameters while resulting in reasonable estimations. We empirically find that placing \modelname~at a shallow-medium layer of the network yields better results (See Section~\ref{sec:ablation}), which is a trade-off between model capacity for estimation and 3D feature encoding.

\subsection{3D Estimation of the Input Tokens}
\modelname~first estimates both the camera-centered 3D coordinates of each token using depth estimation and a camera matrix shared by all the tokens of an image. Then, the camera-centered 3D coordinates are transformed to the ``world'' coordinates by the camera matrix. 


\paragraph{Pseudo-depth Estimation.}
Given input tokens $S=\{s_1, \dots, s_N\}\in \mathbb{R}^{N\times m}$, \modelname~first performs pseudo-depth estimation of each token $s_n$.
The pseudo-depth estimator is a function $f: \mathbb{R}^m \to \mathbb{R}$ that outputs the depth $d_n=f(s_n)$ of each token individually.
We implement $f$ using a 2-layer MLP.
We call this pseudo-depth estimation since it is similar to depth estimation from monocular images but operates at a very coarse scale, given that each token corresponds to an image patch rather than a single pixel in Transformer. 

After pseudo-depth estimation, \modelname~transforms the pseudo-depth map to camera-centered 3D coordinates.
Recall that in ViT~\cite{dosovitskiy2020vit}, an image $X$ is decomposed into $N$ patches $\{X_1, \dots, X_N\} \in \mathbb{R}^{N \times P \times P \times 3}$, where $P\times P$ is the size of each image patch and the tokens $S$ are obtained from a linear projection of these image patches. 
Thus, each token is initially associated with a 2D location on the image plane, denoted as $(u,v)$.
By depth estimation, \modelname~associates each token with one more value $d$.
Based on the pinhole camera model explained in Section~\ref{background}, \modelname~transforms $u,v,d$ to a camera-centered 3D coordinate $(x,y,z)$ by:
\begin{align}
    p^{\text{cam}}_n = 
    \begin{bmatrix} x_n\\ y_n\\ z_n
    \end{bmatrix} = 
    \begin{bmatrix} u_nz_n/c \\ v_nz_n/c \\ z_n
    \end{bmatrix},~
    \text{where} ~
    z_n = d_n.
\end{align}
Since we purely perform the aforementioned estimation from monocular images and the camera intrinsic matrix is unknown, we simply set $c$ to a constant hyperparameter. 
To define coordinate system of 2D image plane $(u, v)$, we set the center of the original image is the origin $(0, 0)$ for convenience, so that the image plane and camera coordinate system shares the same origin. 
We use the center of the image patch to represent its associate $(u, v)$ coordinates.

We believe that this depth-based 3D coordinate estimation best leverages the known 2D geometry, which is beneficial for later representation learning. 
We later confirm this in our ablation study (in Section~\ref{sec:insert_module}), where we compare it against a variant of directly estimating $(x,y,z)$ instead of depth.

\paragraph{Camera Parameter Estimation.}
The camera parameters are required to transform the estimated camera-centered 3D coordinates $p^{\text{cam}}$ to the world coordinate system. These camera parameters are estimated jointly from all input tokens $S$. This involves estimation of two matrices, a $3\times 3$ rotation matrix $\mathbf{R}$ and a $3 \times 1$ and translation matrix $\mathbf{t}$ through an estimator $g$. We implement $g$ using a MLP. Specifically, we use a shared MLP stem to aggregate all the tokens into an intermediate representation. Then, we use two separated fully connected heads to estimate the parameters in $\mathbf{R}$ and $\mathbf{t}$ respectively. We note the camera parameter estimator as a whole:
    $[~\mathbf{R}|\mathbf{t}~] = g(S)$.
To ensure $\mathbf{R}$ is mathematically valid, we first estimate the three values corresponding to yaw, pitch and roll angles of the camera pose, and then convert them into a $3\times 3$ rotation matrix. Research has shown that the discontinuity occurs at the boundary cases in rotation representation~\cite{zhou2019continuity, xiang2021eliminating}, however, such corner cases are rare.

In case of generic visual understanding tasks like object classifications, we expect the camera parameter estimation to perform an ``object-centric canonicalization'' of images with respect to the ``common'' poses of the class object. This is qualitatively shown by Figure~\ref{fig:cup_multipose} in the Appendix.

\paragraph{Transform to World Coordinates.}
Now, with the camera parameters, \modelname~transforms estimated camera-centered coordinates $p^{\text{cam}}$ into the world space, a 3D space where 3D coordinates of the tokens are absolute and viewpoint-invariant.
Following the pinhole camera model, we recover $p^{\text{world}}$:
\begin{equation}
    p^{\text{world}}_n = [~\mathbf{R}^T|\mathbf{R}^T\mathbf{t}~]~p^{\text{cam}}_n.
\end{equation}

\subsection{Incorporating 3D Positional Information in Transformers}
The last step of \modelname~is to leverage the estimated 3D positional information in Transformer backbone. For this, we choose to adopt a typical technique of incorporating positional embedding that is already used in Transformers~\cite{dosovitskiy2020vit,attention}.
In contrast to 2D positional embedding in ViTs~\cite{dosovitskiy2020vit}, \modelname~learns a 3D embedding function $h: \mathbb{R}^{3} \to \mathbb{R}^{m}$ to transform estimated world coordinates $p^{\text{world}}$ to positional embeddings $p^{\text{3D}}$. This 3D embedding function $h$ is implemented using a two-layer MLP. Then, the obtained 3D positional embedding is incorporated in the Transformer backbone by combining it with the token representations. The outcome is the final token representations $\{s^{\text{3D}}\}$:
\begin{align} \label{incorporate}
    s^{\text{3D}}_n = s_n + p^{\text{3D}}_n,\; \text{where}\; p^{\text{3D}} = h(p^{\text{world}}).
\end{align}
After 3D embedding, the resultant token representations are associated with a 3D space, thus enabling the remaining Transformer layers to encode viewpoint-agnostic token representations. We ablate other ways of incorporating the 3D positional information of the tokens in Section~\ref{sec:ablation}.

\subsection{\modelname~in Video Models} \label{video_npl}
Notably, \modelname~can be also easily generalized to video models. 
For video models, the input to \modelname~is a set of spatial-temporal tokens $\{S_1, \dots, S_T\}$ corresponding to a video clip containing $T$ frames, where $S_t=\{s_{t1}, \dots, s_{tN}\}$ are $N$ tokens from $t$-th frame.
We simply extend our module to operate on an additional time dimension, where
depth estimation and 3D positional embedding are done for each spatial-temporal tokens $s_{tn}$ individually: $d_{tn} = f(s_{tn}),\; p^{\text{3D}}_{tn} = h(p^{\text{world}}_{tn})$.
Camera parameters are estimated per input frame ($S_t$) in a dissociated manner, namely $[~\mathbf{R}|\mathbf{t}~]_t=g(S_t)$, resulting in a total of $T$ camera matrices per video.
We investigate another strategy of camera estimation in the supplementary, where only one camera matrix is learned for all frames. However, our studies have substantiated the effectiveness of learning dissociated camera matrices per frame.

\section{Experiments}
We conduct extensive experiments to demonstrate the efficacy of viewpoint-agnostic representations learned by \modelname~in multiple vision tasks: (i) image classification, (ii) multi-view video alignment, and (iii) video action classification. We also qualitatively evaluate the pseudo-depth and camera estimation to confirm \modelname~works.

\subsection{Image Classification}\label{sec:image_classification}
In order to validate the power of \modelname, we first evaluate ViT~\cite{dosovitskiy2020vit} with \modelname~for image classification task using CIFAR-10~\cite{cifar}, CIFAR-100~\cite{cifar} and ImageNet-1K~\cite{imagenet} datasets. 
\paragraph{Training} We use the training recipe of DeiT~\cite{deit} for training our baseline Vision Transformer model on CIFAR and ImageNet datasets from \textit{scratch}. We performed ablations to find an optimal location in ViTs where \modelname~should be plugged-in (Section~\ref{sec:insert_module}). Thus, in all our experiments we place \modelname~after 4 Transformer layers, unless otherwise stated. The configuration of our DeiT-T, DeiT-S, and DeiT-B is identical to that mentioned in~\cite{deit}. All our transformer models are trained for 50 and 300 epochs for CIFAR and ImageNet respectively.
Further training details and hyper-parameters can be found in the supplementary material.
\begin{figure}[thbp]
    \begin{minipage}{0.63\textwidth}
        \centering
        \setlength{\tabcolsep}{3pt}
        \captionof{table}{Top 1 classification accuracy (\%) on CIFAR-10 and 100, ImageNet-1K (IN-1K), viewpoint-perturbed IN-1K (IN-1K-\textit{p}), and ObjectNet. We also report the number of parameters (\#params) and computation (in MACs). Note that the MACs are reported w.r.t. IN-1K samples.} 
        \resizebox{\textwidth}{!}{
        \begin{tabular}{l|cc| l l l l l }
            \toprule
            \textbf{Method} & \textbf{\#params} & \textbf{MACs} & \textbf{CIFAR-10} & \textbf{CIFAR-100} & \textbf{IN-1K} & \textbf{IN-1K-\textit{p}}  & \textbf{ObjectNet}\\
            \midrule
            DeiT-T & 5.72M & 1.08G & 74.1 & 51.3 & 73.4 & 61.3 & 21.3\\ 
              ~\textbf{+\modelname} & 5.95M & 1.10G & \textbf{78.8}\perfinc{4.7} & \textbf{53.7}\perfinc{2.4} & \textbf{73.6}\perfinc{0.2} & \textbf{64.6}\perfinc{3.3} & \textbf{22.4}\perfinc{1.1}\\
            \cmidrule{1-8}
            DeiT-S & 22.1M & 4.24G & 77.2 & 54.6 & 79.4 & 71.1 & 25.8\\
              ~\textbf{+\modelname} & 23.0M & 4.33G & \textbf{80.7}\perfinc{3.5} &  \textbf{61.5}\perfinc{6.9} & \textbf{79.7}\perfinc{0.3} & \textbf{72.7}\perfinc{1.6} & \textbf{27.1}\perfinc{1.3}\\
            \cmidrule{1-8}
            DeiT-B & 86.6M & 16.7G & 76.6 & 51.9 & 81.0 & 70.6 & 27.0\\
              ~\textbf{+\modelname} & 90.1M & 17.2G & \textbf{82.8}\perfinc{6.2} & \textbf{61.8}\perfinc{9.9} & \textbf{81.2}\perfinc{0.2} & \textbf{74.7}\perfinc{4.1} & \textbf{27.3}\perfinc{0.3} \\
            \bottomrule
        \end{tabular}
        }
        \label{tab:img_cls_result}
    \end{minipage}
    \hfill
    \begin{minipage}{0.35\textwidth}
        \includegraphics[width=\textwidth]{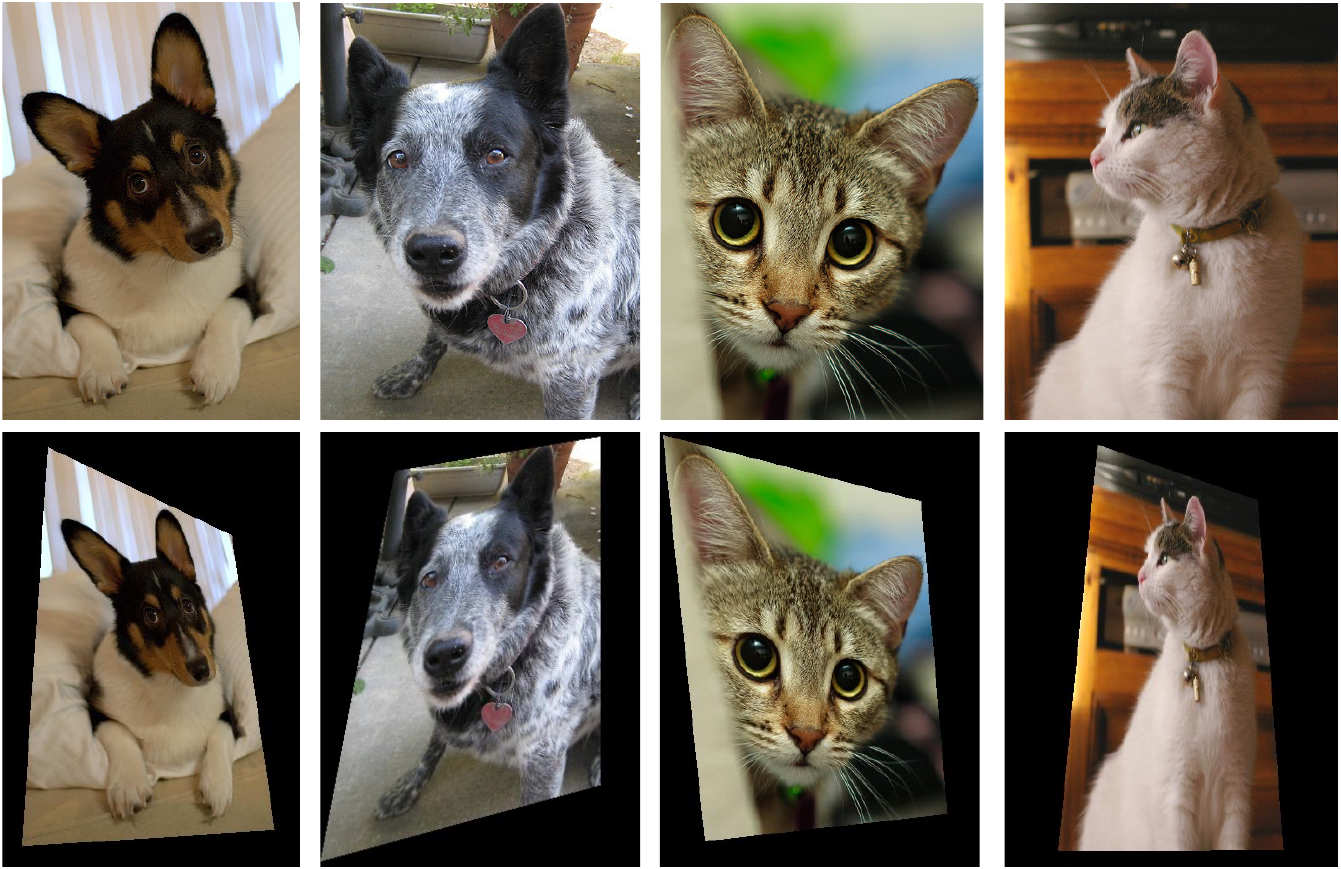}
        \captionof{figure}{\textbf{Top}: original IN-1K samples. \textbf{Bottom}: viewpoint-perturbed IN-1K samples.}
        \label{fig:perturbed}
    \end{minipage}
\end{figure}

\paragraph{Results} From Table~\ref{tab:img_cls_result}, \modelname~with all DeiT variants shows consistent performance improvement over the baseline on the CIFAR datasets, with only $\sim$2\% computation overhead and $\sim$4\% more parameters. Despite fewer training samples, \modelname~significantly outperforms the baseline in CIFAR, showing the strong generalizability of \modelname~when limited training data are available. 
We argue that multi-view data is not available in abundance, especially in domains with limited data (like in medical domain), thus \modelname~with its ability to learn viewpoint-agnostic representations will be crucial in such domains.
We also find the performance improvement on ImageNet is less than that on CIFAR. This is because ImageNet has limited viewpoints in both training and validation splits, thus reducing the significance of performing geometric aware transformations for learning view agnostic representations. 
In contrast, CIFAR samples present more diverse camera viewpoints, so it is a more suitable dataset for testing the quality of learned viewpoint-agnostic representations.

\paragraph{Robustness on Data with Viewpoint Changes}
In order to emphasize the need of learning viewpoint-agnostic representations, we further test the models trained on IN-1K on two test datasets: ObjectNet~\cite{barbu2019objectnet} and ImageNet-1K-perturbed (IN-1K-\textit{p}). ObjectNet~\cite{barbu2019objectnet} is a \textit{test} set designed to introduce more rotation, viewpoint, and background variances in samples compared to ImageNet. Overall, ObjectNet is a very challenging dataset considering large variance in real-world distributions. IN-1K-\textit{p} is a \textit{viewpoint}-perturbed IN-1K validation set by applying random perspective transformations to images, constructed by our own. Example images are shown in Figure~\ref{fig:perturbed}. We note that perspective transformation on these static images is not equivalent to real viewpoint changes. Nonetheless, it is a meaningful for a proof-of-concept experiment.
Results in Table~\ref{tab:img_cls_result} show that models with \modelname~consistently outperform their corresponding Transformer baselines in these two test sets, suggesting \modelname~is trained to generalize across viewpoint changes.

\subsection{Multi-view Video Alignment}\label{sec:video_alignment}
Video alignment~\cite{hadji2021representation,dwibedi2019temporal,misra2016shuffle} is a task to learn a frame-to-frame mapping between video pairs with close semantic embeddings. 
In particular, we consider a \textit{multi-view} setting that aligns videos captured from the same event but different viewpoints, which could further facilitate robot imitation learning from third-person views~\cite{Stadie2017ThirdPersonIL,sermanet2018time,shang2021disentangle,triton}. Here, video pairs from the same event are temporally synchronized. 
\begin{figure}[tbhp]
    \centering
    \scalebox{0.9}{
    \includegraphics[width=\textwidth]{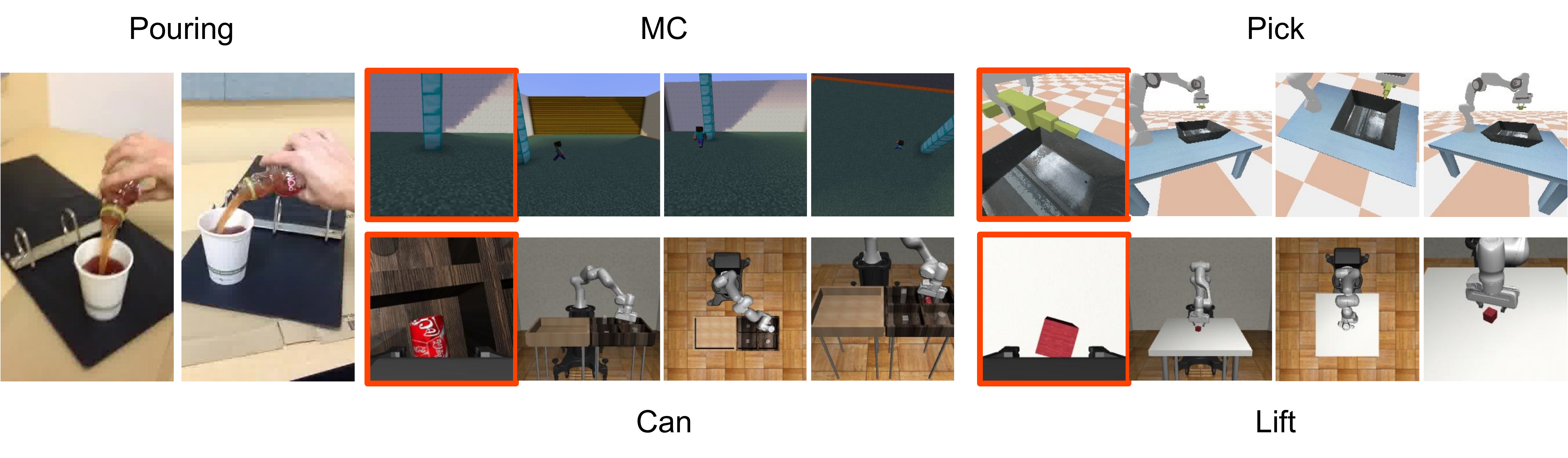}
    }
    \caption{Examples video alignment datasets. Each dataset has synchronized videos of at least 2 viewpoints. All datasets except Pouring have one ego-centric view, highlighted in red boxes. 
    More details are available in supplementary material.}
    \label{fig:align_datasets}
\end{figure}

\paragraph{Datasets} We use 5 multi-view datasets from a wide range of environments: \textbf{Minecraft} (MC) -- video game, \textbf{Pick}, \textbf{Can}, and \textbf{Lift} from robot simulators (PyBullet~\cite{coumans2019} and Robomimic~\cite{robomimic2021}), and \textbf{Pouring} from real-world human actions~\cite{sermanet2018time}. Example video frames are provided in Figure~\ref{fig:align_datasets}. 
Each dataset contains synchronized videos from multiple cameras (viewpoints). There is one ego-centric camera per dataset except Pouring, which are continuously moving with the subject. These ego-centric videos make the alignment more challenging.
Detail dataset statistics are available in supplementary.
\paragraph{Training}
We follow common video alignment methods~\cite{sermanet2018time} to train an encoder that outputs frame-wise embeddings.
We still use DeiT~\cite{deit} as a baseline model (DeiT+TCN) and apply \modelname~to it (+\modelname), similar to image classification.
During training, we use the time-contrastive loss~\cite{sermanet2018time} to encourage temporally closed embeddings to be similar while temporally far-away embeddings to be apart.
Then, we obtain alignments via nearest-neighbor such that an embedding $u_{i}$ from video 1 is being paired to its nearest neighbor $v_{j}$ in video 2 in the embedding space. And similarly $u_{j}$ from video 2 is paired with its nearest neighbor $v_{k}$ in video 1. 
We use ImageNet-1K pre-trained weights for experiments on Pouring, but we train from scratch for other datasets considering that simulation environments are out of real-world distribution.

\paragraph{Evaluation} We evaluate the alignment by three metrics: Alignment Error~\cite{sermanet2018time}, Cycle Error, and Kendall's Tau~\cite{kendall1938new}.
Let the alignment pairs between two videos be $(u_i, v_j)$ and $(v_j, u_k)$.
In brief, Alignment Error measures the temporal mismatching $|i-j|$ of $(u_i,v_j)$. Cycle Error is based on cycle-consistency~\cite{wang2019learning, dwibedi2019temporal}, where two pairs $(u_i, v_j)$ and $(v_j,u_k)$ are called \textit{consistent} when $i = k$. Thus, Cycle Error measures the inconsistency based on distance metric $|i-k|$. Kendall's Tau ($\tau$) measures ordering in pairs. Given a pair of embeddings from video 1 $(u_i, u_j)$ and their corresponding nearest neighbors from video 2 $(v_p, v_q)$, the indices tuple $(i,j,p,q)$ is \textit{concordant} when $i < j$ and $p < q$ or $i > j$ and $p > q$. Otherwise the tuple is \textit{discordant}. Kendall's Tau computes the ratio of concordant pairs and discordant pairs over all pairs of frames. Let $N$ be the number of frames in a video, then the formal notations of the three metrics are:
\begin{align}
\resizebox{0.85\textwidth}{!}{
    $\text{\footnotesize Alignment Error} = \mathbb{E}_i\frac{|i-j|}{N};\;
    \text{\footnotesize Cycle Error} = \mathbb{E}_i\frac{|i-k|}{N};\;
    \tau = \frac{\text{\# concordant pairs}-\text{\# discordant pairs}}{N(N-1)/2}$.
}
\end{align}
We establish two evaluation protocols: (a) \textbf{Seen} and (b) \textbf{Unseen}. In \textbf{Seen}, we train and test models on videos from all cameras. However, in \textbf{Unseen}, we hold out several cameras for test,
which is a representative scenario for validating the effectiveness of \modelname. 
Detail of experimental settings are provided in supplementary.

\begin{figure}
    \centering
    \includegraphics[width=\textwidth]{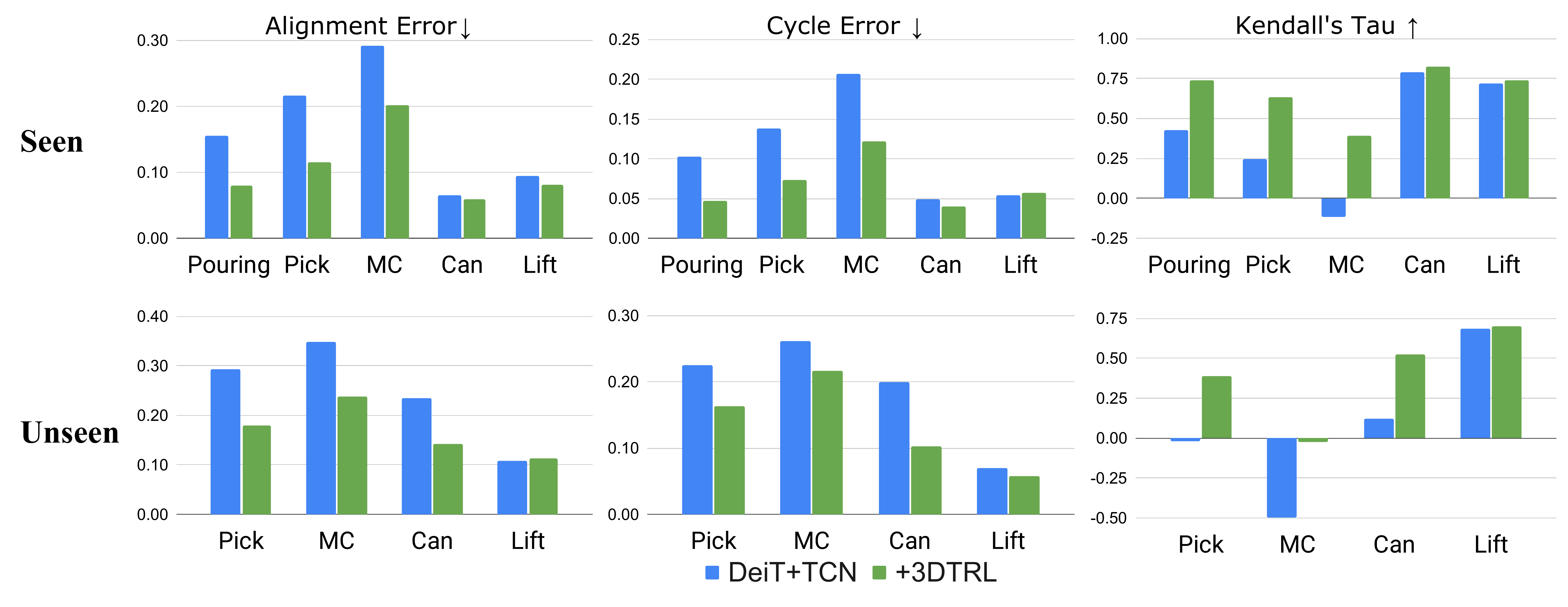}
    \caption{Results on video alignment in \textbf{Seen} and \textbf{Unseen} protocols. $\uparrow$ indicates a higher metric is better and $\downarrow$ is otherwise. Blue bars are for DeiT+TCN without \modelname~and green bars are with \modelname. \modelname~outperforms the baseline consistently in both settings. 
    We note \textbf{Unseen} is not applicable to Pouring dataset because only two cameras are available in this dataset.
    }
    \label{fig:video_align_results}
\end{figure}

\paragraph{Results}
Figure~\ref{fig:video_align_results} illustrates the evaluation results of 2 viewpoint settings over 5 datasets, compared to the DeiT baseline.
\modelname~outperforms the baseline consistently across all datasets. In particular, \modelname~improves the baseline by a large margin in Pouring and MC, corroborating that \modelname~adapts to diverse unseen viewpoints.
The improvements of \modelname~on Pick and MC also suggests the strong generalizability when learning from smaller datasets. With enough data (Lift \& Can), \modelname~still outperforms but the gap is small.
When evaluating in \textbf{Unseen} setting, both methods have performance drop. However, \modelname~still outperforms in Pick, MC, and Can, which suggests the representations learned by \modelname~are able to generalize over novel viewpoints. MC has the largest viewpoint diversity so it is hard to obtain reasonable align results in the unseen setting for both the models. 
\begin{figure}[htbp]
    \begin{minipage}{0.55\textwidth}
        \centering
        \setlength{\tabcolsep}{2pt}
        \captionof{table}{Video alignment results compared with SOTA methods. Values are alignment errors.}
        \resizebox{\textwidth}{!}{
        \begin{tabular}{lccccc}
        \toprule
            \textbf{Method} & \textbf{Backbone} & \textbf{Input} & \textbf{Pouring} & \textbf{Pick} & \textbf{MC} \\
            \midrule
            TCN~\cite{sermanet2018time} & CNN & 1 frame & 0.180 & 0.273 & 0.286 \\
            Disentanglement~\cite{shang2021disentangle} & CNN & 1 frame & - & 0.155 & 0.233 \\ 
            \midrule
            mfTCN~\cite{dwibedi2018learning} & 3DCNN & 8 frames & 0.143 & - & - \\
            mfTCN~\cite{dwibedi2018learning} & 3DCNN & 32 frames & 0.088 & - & -\\
            \midrule
            DeiT~\cite{deit}+TCN & Transformer & 1 frame & 0.155 & 0.216 & 0.292 \\
            ~\textbf{+\modelname} & Transformer & 1 frame & \textbf{0.080} & \textbf{0.116} & \textbf{0.202} \\
        \bottomrule
        \end{tabular}
        }
        \label{tab:video_align_compare_sota}
    \end{minipage}
    \begin{minipage}{0.44\textwidth}
    \includegraphics[width=\textwidth]{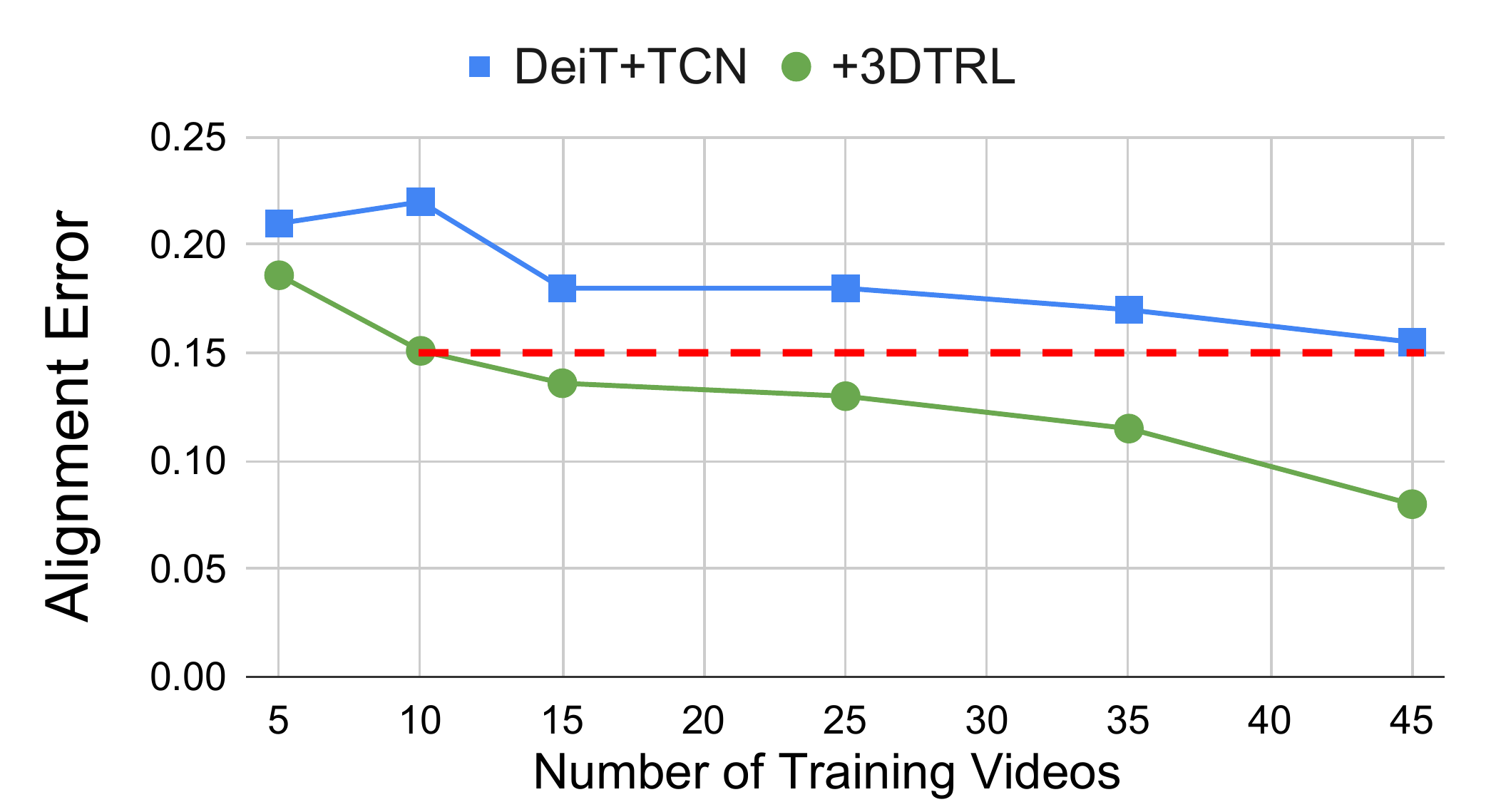}
    \captionof{figure}{\small{Alignment error} w.r.t \# training videos. Red dashed line indicates \modelname~using \small{10 videos} outperforms DeiT using \small{45 videos}.}
    \label{fig:error_train_videos}
    \end{minipage}
\end{figure}

In Table~\ref{tab:video_align_compare_sota}, we further compare \modelname~with previous methods on certain datasets. 
Note that Disentanglement~\cite{shang2021disentangle} uses extra losses whereas others use time-contrastive loss only. We find that \modelname~with only \textit{single-frame} input is able to surpass the strong baselines set by models using extra losses~\cite{shang2021disentangle} or multiple input frames~\cite{dwibedi2018learning}. 
We also vary the number of training videos in Pouring dataset, and results from Figure~\ref{fig:error_train_videos} show that \modelname~benefits from more data. Meanwhile, \modelname~can outperform the baseline while only using 22\% of data the baseline used.

\subsection{Quantitative Evaluations on Recovering 3D information}\label{sec:3d_eval}
We perform quantitative evaluations on how well \modelname~recovers 3D information. We emphasize once more that the 3D recovery in our approach is done without any supervision; it is optimized with respected to the final loss (e.g., object classification), without any access to the ground truth 3D information.
We first discuss 3D estimations focusing on pseudo-depth estimation (Section~\ref{sec:depth_eval}), then we evaluate camera estimation (Section~\ref{sec:camera_eval}). 
\subsubsection{3D Estimation Evaluation}\label{sec:depth_eval}
\begin{figure}[htbp]
    \begin{minipage}{0.32\textwidth}
        \centering
        \includegraphics[width=\linewidth]{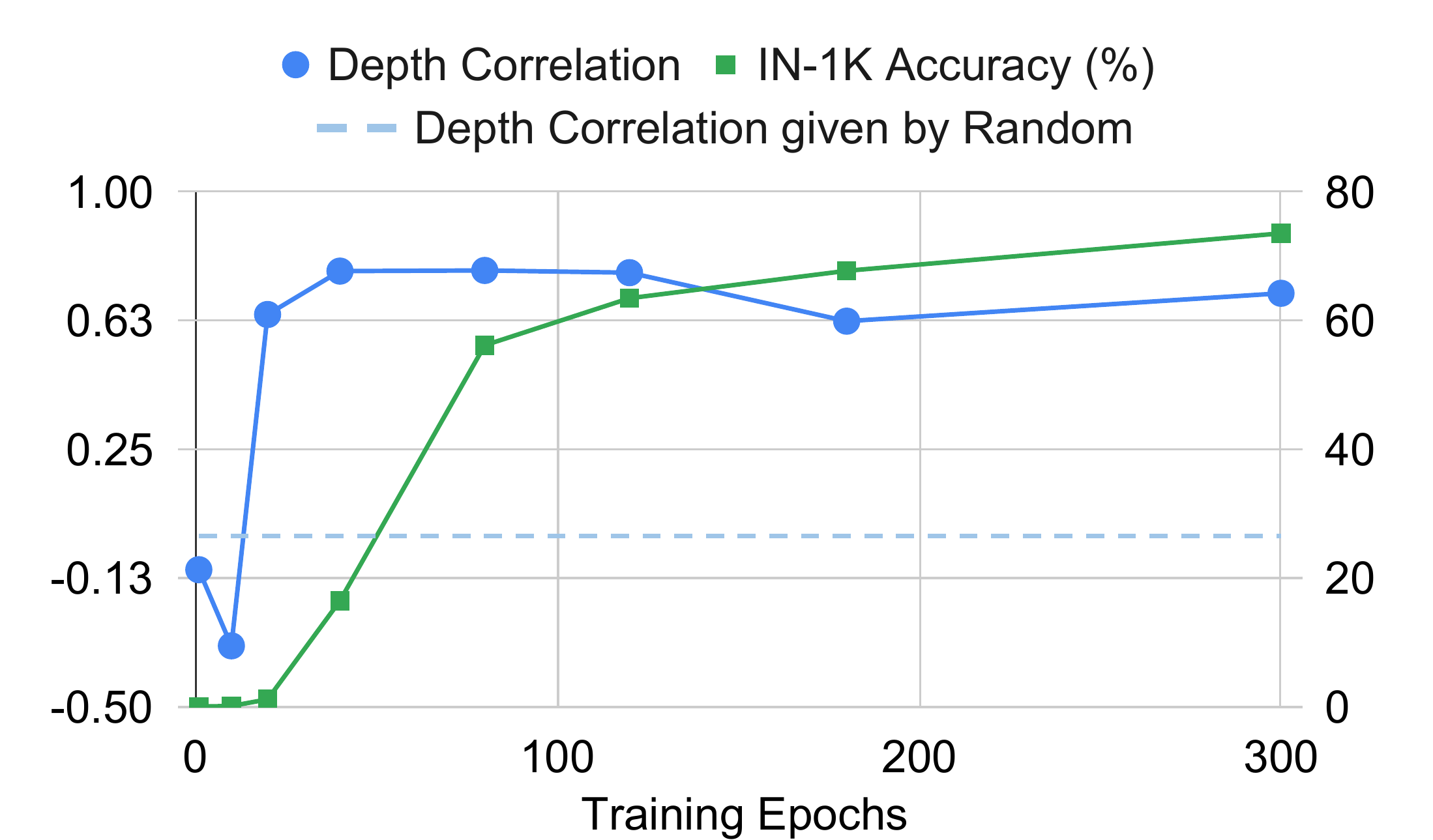}
        \captionof{figure}{Depth correlation evaluation results. We find our estimation has a high ($\sim$0.7) correlation to the ground truth.}
        \label{fig:depth_corr}
    \end{minipage}
    \hfill
    \begin{minipage}{0.65\textwidth}
        \centering
        \includegraphics[width=0.48\linewidth]{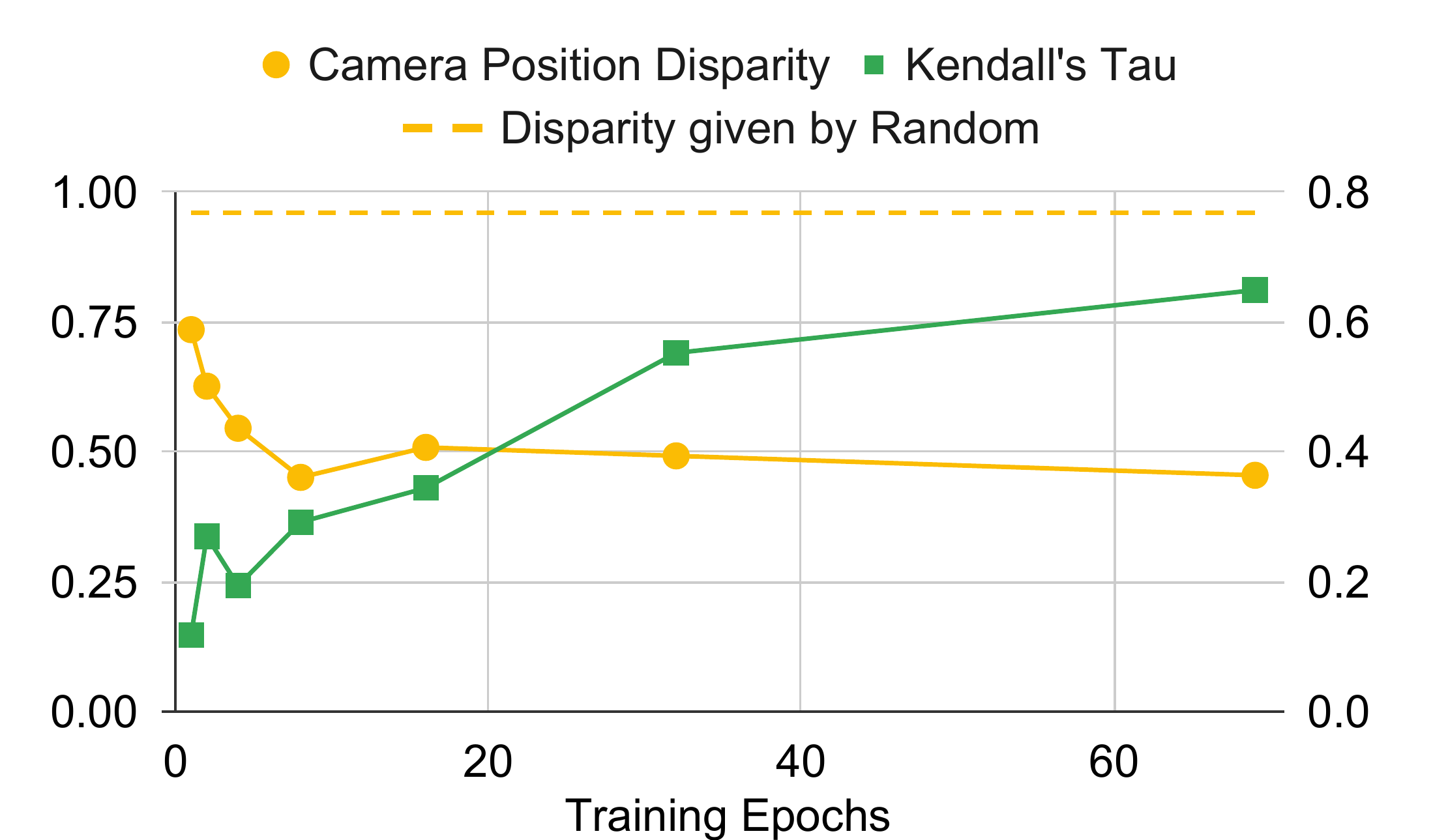}
        \includegraphics[width=0.48\linewidth]{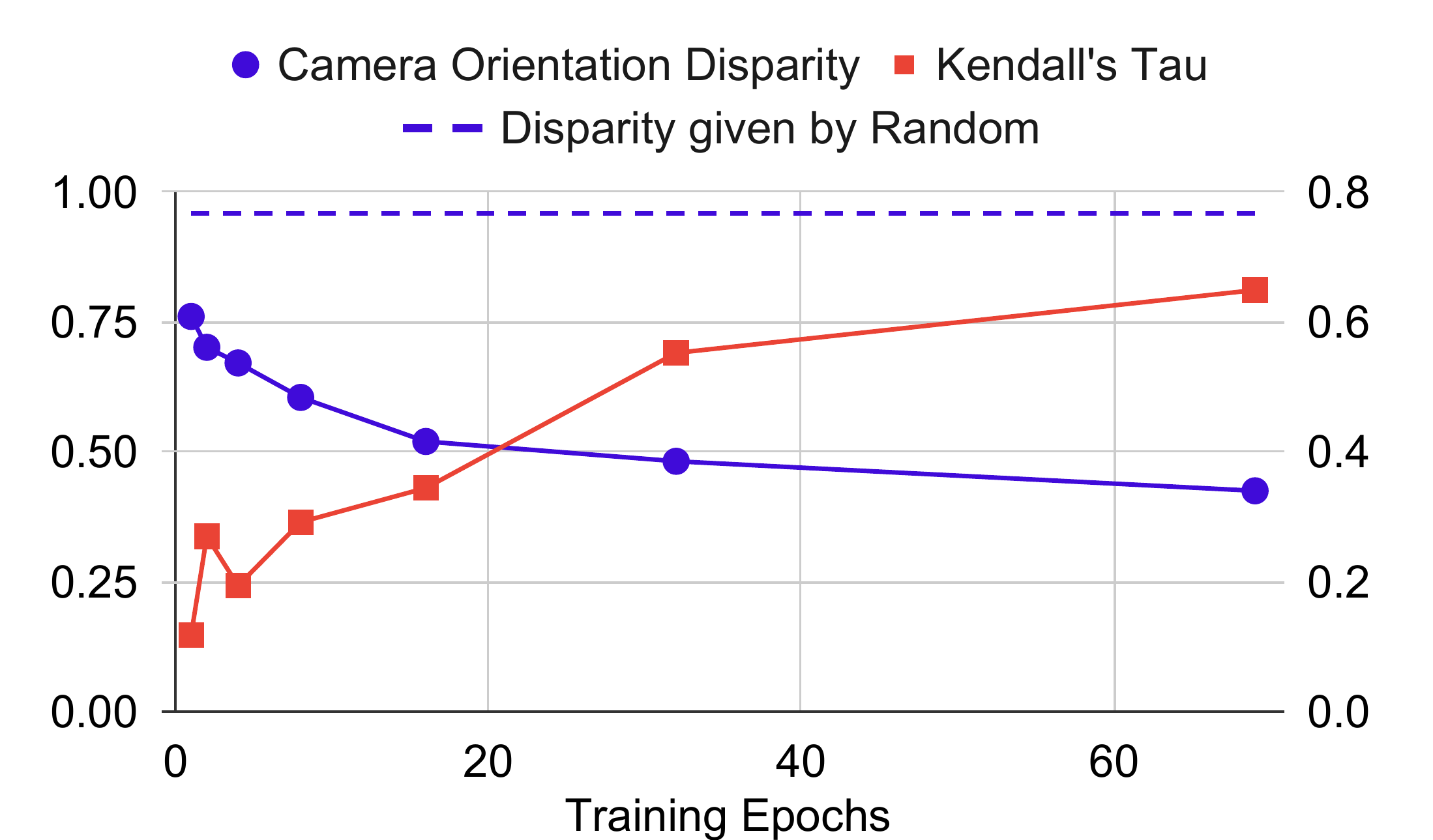}
        \captionof{figure}{Quantitative evaluations on camera position (left) and camera orientation (right) respectively. Overall, we show the estimated cameras from \modelname~have an acceptable mapping to the ground truth (both give $<0.5$ disparity).}
    \label{fig:campos_eval}
    \end{minipage}
\end{figure}

The key component of our 3D estimation is pseudo-depth estimation. In order to evaluate the 3D estimation capability, we compare the pseudo-depth map with ground truth depth map, using NeRF~\cite{mildenhall2020nerf} dataset. 
We test the pseudo-depth with DeiT-T+\modelname~trained on IN-1K.
\paragraph{Metric: Depth Correlation.} Since our estimated pseudo-depth ($d'$) and ground truth ($d$) are in different scales, we measure their relative correspondence, i.e., correlation of two sets of data. We use Pearson’s r: $r = \text{correlation}(d, d')$, where we regard two depth maps as two groups of data. Note that \modelname~operates on visual Transformers, so the pseudo-depth map is at a very coarse scale (14$\times$14). We also resize the given depth map to 14$\times$14 to perform the evaluation. We report the average of $r$ across all evaluation subsets.\footnote{To do so, we convert $r$ to Fisher’s $z$, take the mean value across all subsets, and 
convert back to Pearson’s $r$.}.
\paragraph{Results.} Figure~\ref{fig:depth_corr} shows the evaluation results. 
We find the estimated pseudo-depth highly correlates with the ground truth ($r\approx0.7$).  
Compared to the random prediction baseline, the depth from \modelname~shows much higher correlation to the ground truth. We also find that the model learns the estimated depth with higher correlation in the earlier stage of the training (e.g., 20 epochs) and then drops a bit in the later training epochs ($\sim$ 0.07 drop).
\subsubsection{Camera Estimation Evaluation}\label{sec:camera_eval}
In this evaluation, we mainly answer how well \modelname~estimates camera position and orientation. To do this, we evaluate DeiT+\modelname~trained in previous video-alignment task (Section~\ref{sec:video_alignment}), using \textbf{Can} dataset. Recall that we train the model from scratch for the video alignment, without access to ground truth 3D information. We use first-person view videos for our evaluation --- the camera moves together with the robot and our objective is to estimate its pose. We introduce two metrics.

\textbf{Metric1: Camera Position Disparity}: For each video, we get the estimated camera positions $\{\mathbf{p}'\}$ and ground truth camera positions $\{\mathbf{p}\}$, which are 3-D vectors. Since estimated and ground truth cameras are in two different coordinate systems, we need to measure how estimated camera positions map to the ground truth in a scale, translation and rotation-invariant way. The existing metrics like AUROC~\cite{sarlin2020superglue,jin2021planar} are not applicable. Therefore, we use Procrustes analysis~\cite{gower1975generalized} and report the disparity metric. The disparity value ranges $[0, 1]$, where a lower value indicates that two sets are more similar. We take the average disparity of all videos.

\textbf{Metric2: Camera Orientation Disparity}: Each camera has its orientation, i.e. ``looking-at'' direction, described by a 3-D vector. We note the estimated camera orientations $\{\mathbf{o}'\}$ and ground truth $\{\mathbf{o}\}$. Similar as the camera position disparity mentioned above, we use the disparity given by Procrustes analysis~\cite{gower1975generalized} to measure how $\{\mathbf{o}'\}$ and $\{\mathbf{o}\}$ align. We report the mean disparity from all the videos.
\paragraph{Results.} Results are shown in Figure~\ref{fig:campos_eval}. Both position and orientation disparity of the final model are relatively small ($<0.5$), showing that the camera estimation from \modelname~is partially aligned with the ground truth.
We also observe that the disparity decreases over the training epochs.

\subsection{Qualitative Evaluations}
\begin{figure}
    \centering
    \includegraphics[width=\textwidth]{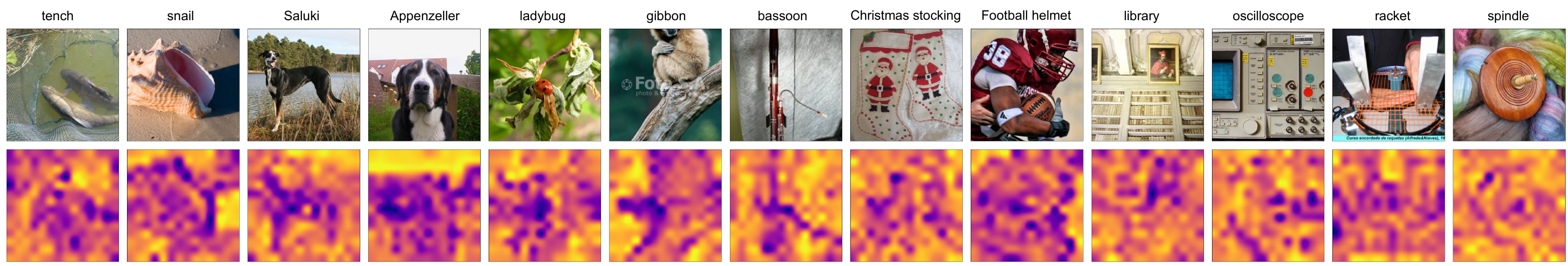}
    \caption{\textbf{Top}: IN-1K samples with corresponding class label. \textbf{Bottom}: Estimated pseudo-depth map, interpolated from $16\times 16$ to $224\times 224$ for better understanding. Depth increases from blueish to yellowish region. More examples are in Figure~\ref{fig:morevisdepth} in Appendix.}
    \label{fig:vis}
\end{figure}

We visualize some qualitative results for an intuitive understanding of \modelname's effectiveness.
The visualization contains estimated pseudo-depth maps which are from the key intermediate step of \modelname. We use ImageNet-1K trained DeiT-T+\modelname~and its validation samples for visualization. In Figure~\ref{fig:vis}, we observe a fine separation of the object-of-interest and the background in most of the pseudo-depth maps. Thus, these predicted pseudo-depth maps are sufficient to recover the tokens corresponding to object-of-interest in 3D space. For a qualitative evaluation on camera pose, please refer to Figure~\ref{fig:cup_multipose} in the Appendix, where we show the images with the same/similar object pose (in different environments) result in similar estimated extrinsics regardless of the background. We believe our estimated extrinsics are doing object-centric canonicalization of images with respect to their object poses, in order to optimize the representations for the final losses.
\subsection{\modelname~on More Transformer Architectures}
\modelname~is designed to be a plug-and-play model for Transformers. We test \modelname~with more Transformer architectures on CIFAR and two multi-view video alignment datasets. We use Swin~\cite{liu2021Swin} (Tiny) and TnT~\cite{tnt} (Small). Results are provided in Table~\ref{tab:more_transformers}.
We find that \modelname~generally improves the performance of two Transformer architectures in all the datasets.
This confirms \modelname~is applicable to different Transformer architectures. The relative small improvement over the Swin backbone is due to the strong inductive bias from its local window. 
\begin{table}[htbp]
    \centering
    \caption{Results of using \modelname~in more Transformer architectures, on CIFAR and multi-view video alignment datasets. Reported numbers are accuracy for CIFAR and Kendall's tau for video alignment, both are the higher the better. We show \modelname~generally improves the performance in all tasks.}
    \resizebox{0.6\textwidth}{!}{
    \begin{tabular}{lllll}
    \toprule
    \textbf{Model}     & \textbf{CIFAR-10} & \textbf{CIFAR-100} & \textbf{Pouring} & \textbf{Pick} \\
    \midrule
    Swin-T     & 50.11 & 21.53 & 0.584 & 0.623 \\
    ~\textbf{+\modelname} & \textbf{50.29}\perfinc{0.18} & \textbf{21.55}\perfinc{0.02} & \textbf{0.683}\perfinc{0.099} & \textbf{0.640}\perfinc{0.017}\\
    \midrule
    TnT-S & 81.25 & 54.07 & 0.740 & 0.640 \\
    ~\textbf{+\modelname} & \textbf{82.43}\perfinc{1.18} & \textbf{56.00}\perfinc{1.93} & \textbf{0.792}\perfinc{0.052} & \textbf{0.671}\perfinc{0.031} \\
    \bottomrule
    \end{tabular}
    }
    \label{tab:more_transformers}
\end{table}
\subsection{Ablation Studies}\label{sec:ablation}
We conduct our ablation studies on image models mostly using \textit{CIFAR} for image classification and \textit{Pick} for multi-view video alignment.
\begin{figure}[htbp]
    \vspace{-7pt}
    \begin{minipage}{0.57\textwidth}
        \centering
        \setlength{\tabcolsep}{1pt}
        \captionof{table}{Ablation study results. For CIFAR, we test on models based on tiny (T), small (S) and base (B) backbones (DeiT) and report accuracy(\%). For Pick we only test base model and report alignment error.}
        \resizebox{\textwidth}{!}{
        \begin{tabular}{lccc}
        \toprule
            \textbf{Method} & \textbf{CIFAR-10} (T/S/B) & \textbf{CIFAR-100} (T/S/B) & \textbf{Pick}  \\
            \midrule
            DeiT & 74.1/ 77.2 / 76.6 & 51.3 / 54.6 / 51.9 &  0.216 \\
             DeiT + MLP & 74.2 / 77.2 / 76.5 & 47.9 / 54.7 / 53.4 & 0.130 \\
             DeiT + \modelname & 78.8 / 80.7 / 82.8 & \textbf{53.7} / 61.5 / \textbf{61.8} &  \textbf{0.116} \\
            \midrule
            Depth Estimation $\rightarrow$ $xyz$ Estimation & 76.7 / 78.2 / 77.4 & 48.3 / 54.1 / 52.6 & 0.134 \\
            Embedding $\rightarrow$ Concat. & \textbf{80.7} / \textbf{83.7} / \textbf{84.9} & 53.4 / \textbf{61.8} / 60.2 &  0.133 \\
         \bottomrule
        \end{tabular}
        }
        \label{tab:mlp_vs_modelname}
    \end{minipage}
    \hfill
    \begin{minipage}{0.42\textwidth}
        \includegraphics[width=\linewidth]{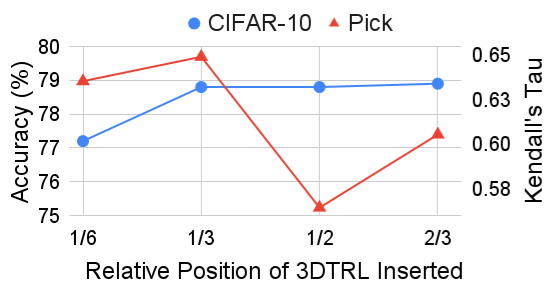} 
        \captionof{figure}{
        Results on inserting \modelname~at different locations.}
        \label{fig:insert_module}
    \end{minipage}
    \vspace{-15pt}
\end{figure}
\paragraph{MLP vs. \modelname.} \modelname~is implemented by several MLPs with required geometric transforms in between. In this experiment, we replace \modelname~with the similar number of fully-connected layers with residual connection, to have comparable parameters and computation as \modelname. Results are provided in Table~\ref{tab:mlp_vs_modelname}. We find that MLP implementation is only comparable with the baseline performance despite the increase in parameters and computation. Thus, we confirm that the geometric transformations imposed on the token representations is the key to make \modelname~effective.
\paragraph{Token Coordinates Estimation.} In this ablation, we show how estimating only depth compared to estimating a set of 3 coordinates $xyz$ differs in \modelname. We find at Line 4 and 5 in Table~\ref{tab:mlp_vs_modelname} that depth estimation is better because it uses precise 2D image coordinates when recovering tokens in 3D space, whereas estimating $xyz$ regresses for the 3D coordinates without any geometric constraints. Also, we find that estimating $xyz$ hampers the performance in image classification task more than video alignment. This is because estimating 3D coordinates is harder when the training samples are in unconstrained scenarios. In contrast, video pairs in an alignment dataset share the same scene captured from different view angles, which facilitates the recovery of token in 3D space.
\paragraph{How to incorporate 3D positional information in Transformers?}
By default, we use Equation~\ref{incorporate} to incorporate the 3D positional information in Transformer backbone through a learned positional embedding $p^{3D}$. In this experiment, we directly infuse the estimated 3D world coordinates $p^{\text{world}}$ within the token representation by concatenating them across the channel axis. The $(m+3)$-d feature is then projected back to $m$-d by a MLP. We keep parameters and computation comparable to the default \modelname.
We test this \modelname~variant (Embedding $\rightarrow$ Concat.) and results are presented in Table~\ref{tab:mlp_vs_modelname}. We find that the concatenation variant outperforms the default variant of \modelname~on CIFAR-10, but comparable and worse results in CIFAR-100 and Pick. This observation substantiates the instability of using raw 3D coordinates. In comparison, the use of 3D positional embedding is generalizable to more challenging and diverse scenarios. 

\vspace{-7pt}
\paragraph{Comparison with perspective augmentation} We compare \modelname~with vanilla data augmentation on perspective transforms on video alignment task and results are in Appendix~\ref{sec:perspective_aug}. We confirm \modelname~does better than applying perspective augmentations.

\vspace{-7pt}
\paragraph{Where should we have \modelname?}\label{sec:insert_module}
We vary the location of \modelname~to empirically study the optimal location of \modelname~in a 12-layer DeiT-T backbone. In Figure~\ref{fig:insert_module}, we find that inserting \modelname~at the earlier layers (after 1/3 of the network) yields the best performance consistently on both datasets.

\subsection{\modelname~for Video Representation Learning}
\begin{table}[b]
    \centering
    \caption{Results on action recognition on Smarthome and NTU. \textit{Acc} is classification accuracy (\%) and \textit{mPA} is mean per-class accuracy. 
    In methods using Kinetics-400 (K400) pre-training, TimeSformer backbone is always initialized with pre-trained weights, and \modelname~w/, w/o K400 denotes \modelname~is randomly initialized and is initialized from pre-trained weights respectively.}
    \resizebox{0.75\textwidth}{!}{
    \begin{tabular}{llllll}
        \toprule
        \multirow{2}{*}{\textbf{Method}} & \multicolumn{2}{c}{\textbf{Smarthome} (CV2)}  & \multicolumn{2}{c}{\textbf{Smarthome} (CS)} & \textbf{NTU} (CV) \\ 
        \cmidrule(lr){2-3} \cmidrule(lr){4-5} \cmidrule(lr){6-6}
         &  \textit{Acc} & \textit{mPA} & \textit{Acc} & \textit{mPA} & \textit{Acc} \\
         \midrule
         TimeSformer~\cite{timesformer}  & 59.4 & 27.5 & 75.7 & 56.1 & 86.4 \\
         + \textbf{\modelname }& \textbf{62.9}\perfinc{3.5} & \textbf{34.0}\perfinc{6.5} & \textbf{76.1}\perfinc{0.4}  & \textbf{57.0}\perfinc{0.9} & \textbf{87.9}\perfinc{1.5} \\
         \midrule
         \multicolumn{6}{@{}l}{\makecell{\hspace{1.3in}\underline{\textbf{ Kinetics-400 pre-trained}}}} \vspace{0.1in}\\ 
         TimeSformer~\cite{timesformer}  & 69.3 & 37.5 & 77.2 & 57.7 & 87.7 \\
         + \modelname~w/o K400  &  69.5\perfinc{0.2} & 39.2\perfinc{1.7} & 77.5\perfinc{0.3} & 58.9\perfinc{1.2} & \textbf{88.8}\perfinc{1.1} \\
          \textbf{+ \modelname~w/ K400} & \textbf{71.9}\perfinc{2.6} & \textbf{41.7}\perfinc{4.2} & \textbf{77.8}\perfinc{0.6} & \textbf{61.0}\perfinc{2.3} & 88.6\perfinc{0.9}\\
         \bottomrule
    \end{tabular}
    }
    \label{tab:action_recognition_results}
\end{table}
In this section, we illustrate how \modelname~can be adapted for video models. Since we aim at learning viewpoint-agnostic representations, as a natural choice we validate the effectiveness of \modelname~on video datasets with multi-camera setups and cross-view evaluation. Consequently, we conduct our experiments on two multi-view action recognition datasets:  \textbf{Toyota Smarthome}~\cite{smarthome} (Smarthome) and \textbf{NTU-RGB+D}~\cite{NTU_RGB+D} (NTU) for the task of action classification. For evaluation on Smarthome, we follow Cross-View 2 (CV2) and Cross-Subject (CS) protocols proposed in~\cite{smarthome}, whereas on NTU, we follow Cross-View (CV) protocol proposed in~\cite{NTU_RGB+D}. In cross-view protocols, the model is trained on a set of cameras and tested on a different set of cameras. Similarly for cross-subject protocol, the model is trained and tested on different set of subjects. 
More details on these datasets are provided in Appendix~\ref{sec:video_exp_setting}. 

\paragraph{Network architecture \& Training / Testing} TimeSformer~\cite{timesformer} is a straightforward extension of ViT~\cite{dosovitskiy2020vit} for videos which operates on spatio-temporal tokens from videos, so that \modelname~can be easily deployed to TimeSformer as well. Similar to our previous experimental settings, we place \modelname~after 4 Transformer blocks in TimeSformer. Please refer to Appendix~\ref{sec:video_exp_setting} for detailed settings.

\paragraph{Results} In Table~\ref{tab:action_recognition_results}, we present the action classification results on Smarthome and NTU datasets with \modelname~plugged in TimeSformer. 

\modelname~can easily take advantage of pre-trained weights because it does not change the relying backbone Transformer -- just being added in between blocks. In Table~\ref{tab:action_recognition_results}, we present results for two fine-tuning scenarios: (a) \modelname~w/o K400  and (b) \modelname~w/ K400. For the first scenario (a), TimeSformer is initialized with K400 pre-training weights and leave the parameters in \modelname~randomly initialized. Then in the fine-tuning stage, all the model parameters including those of \modelname~is trained. In the second scenario (b), all parameters in TimeSformer and \modelname~are pre-trained on K400 from scratch and fine-tuned on the respective datasets. 

We find that all the variants of \modelname~outperforms the baseline TimeSformer results. 
Our experiments show that although there is an improvement with \modelname~compared to the baseline for different fine-tuning strategy, it is more significant when \modelname~is pre-trained with K400. However, when large-scale training samples are available (NTU), \modelname~does not require K400 pre-training. To sum up, \modelname~can be seen as a crucial ingredient for learning viewpoint-agnostic video representations.

\section{Related Work}
There has been a remarkable progress in visual understanding with the shift from the use of CNNs~\cite{lecun1995convolutional,vgg16,resnet,inception} to visual Transformers~\cite{dosovitskiy2020vit}.
Transformers have shown substantial improvements over CNNs in image~\cite{dosovitskiy2020vit,liu2021Swin,crossvit,kahatapitiya2021swat,deit,tnt,yuan2021tokens} analysis and video~\cite{timesformer,svn,ryoo2021tokenlearner,arnab2021vivit,liu2021videoswin} understanding tasks due to its flexibility in learning global relations among visual tokens. 
Studies also combine CNNs with Transformer architectures to leverage the pros in both the structures~\cite{dai2021coatnet, starformer,liu2022convnet,graham2021levit, starformerj}. In addition, Transformer has been shown to be effective in learning 3D representation~\cite{pointtransformer}.
However, these advancements in architecture-types have not addressed the issue of learning viewpoint-agnostic representation. Viewpoint-agnostic representation learning is drawing increasing attention in the vision community due to its wide range of downstream applications like 3D object-detection~\cite{rukhovich2022imvoxelnet},video alignment~\cite{dwibedi2019temporal, gao2022fine, chen2022frame}, action recognition~\cite{sigurdsson2018charades, sigurdsson2018actor}, pose estimation~\cite{haque2016towards,sun2020view}, robot learning~\cite{sermanet2018time,shang2021disentangle,hsu2022vision,jangir2022look,Stadie2017ThirdPersonIL, triton}, and other tasks.

There is a broad line of work towards directly utilizing 3D information like depth~\cite{haque2016towards}, pose~\cite{das2020vpn, das2021vpn++}, and point clouds~\cite{piergiovanni20214d,robert2022learning}, or in some cases deriving 3D structure from paired 2D inputs~\cite{vijayanarasimhan2017sfm}. However, methods rely on the availability of multi-modal data which is hard to acquire are not scalable.

Consequently, other studies have focused on learning 3D perception of the input visual signal in order to generalize the learned representation to novel viewpoints.  This is done by imposing explicit geometric transform operations in CNNs~\cite{stn,NPL_2021_CVPR,yan2016perspective,cao2021monoscene,rukhovich2022imvoxelnet}, without the requirement of any 3D supervision.
In contrast to these existing works, our Transformer-based \modelname~imposes geometric transformations on visual tokens to recover their representation in a 3D space. To the best of our knowledge, \modelname~is the first of its kind to learn a 3D positional embedding associated with the visual tokens for viewpoint-agnostic representation learning in different image and video tasks.

\section{Conclusion}
In this work, we have presented \modelname, a plug-and-play module for visual Transformer that leverages 3D geometric information to learn viewpoint-agnostic representations. Within \modelname, by pseudo-depth estimation and learned camera parameters, it manages to recover positional information of tokens in a 3D space. 
Through our extensive experiment, we confirm \modelname~is generally effective in a variety of visual understanding tasks including image classification, multi-view video alignment, and cross-view action recognition, by adding minimum parameters and computation overhead.
\section*{Acknowledgment}
We thank insightful discussions with members of Robotics Lab at Stony Brook.
This work is supported by Institute of Information \& communications Technology Planning \& Evaluation (IITP) grant funded by the Ministry of Science and ICT (No.2018-0-00205, Development of Core Technology of Robot Task-Intelligence for Improvement of Labor Condition. This work is also supported by the National Science Foundation (IIS-2104404 and CNS-2104416).

\bibliography{refs.bib}
\bibliographystyle{abbrvnat}

\newpage
\appendix

\section{Qualitative Evaluation on Camera Estimation}
\begin{figure}[htbp]
    \centering
    \includegraphics[width=0.98\textwidth]{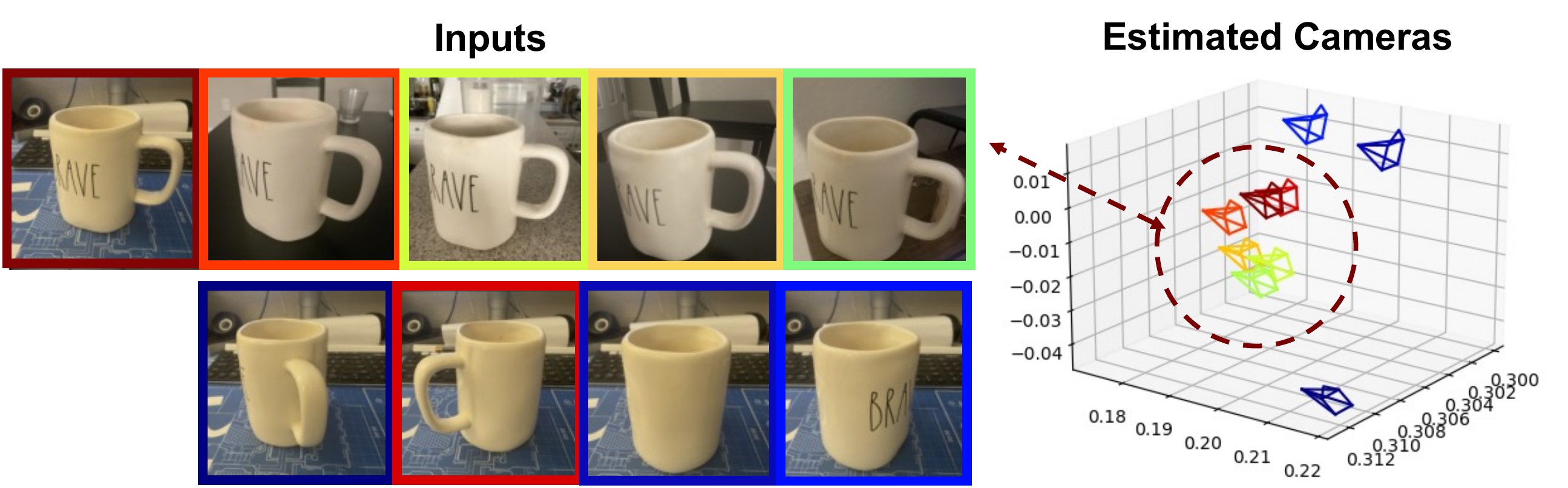}
    \caption{Qualitative experiment on how 3DTRL reacts to similar and different poses of the same object. Photos are taken by a regular smartphone. We use DeiT-T+\modelname~trained on ImageNet. The estimated cameras from the similar poses (first row) are clustered as shown in the red-dashed circle, while the other cameras from different poses (second row) are apart. We notice an outlier whose estimation is also introduced in this cluster, owing to the model invariance to horizontal flipping which is used as an augmentation during training.}
    \label{fig:cup_multipose}
\end{figure}

We collected images of the same object with various object poses and backgrounds, and visualized their camera estimation in Figure~\ref{fig:cup_multipose}. 
This was done with DeiT-T+\modelname~trained on ImageNet. We observe \modelname~estimates similar camera poses (clustered in the red-dashed circle) for similar object poses regardless of different backgrounds. When given different object poses, \modelname~estimates cameras in scattered position and orientations. We notice that there is an outlier whose estimation is also introduced in this cluster, owing to the model invariance to horizontal flipping which is used as an augmentation during training.
This visualization suggests that \modelname, when trained for the object classification task, might be performing object-centric canonicalization of the input images.

\begin{figure}[htbp]
    \centering
    \includegraphics[width=\textwidth]{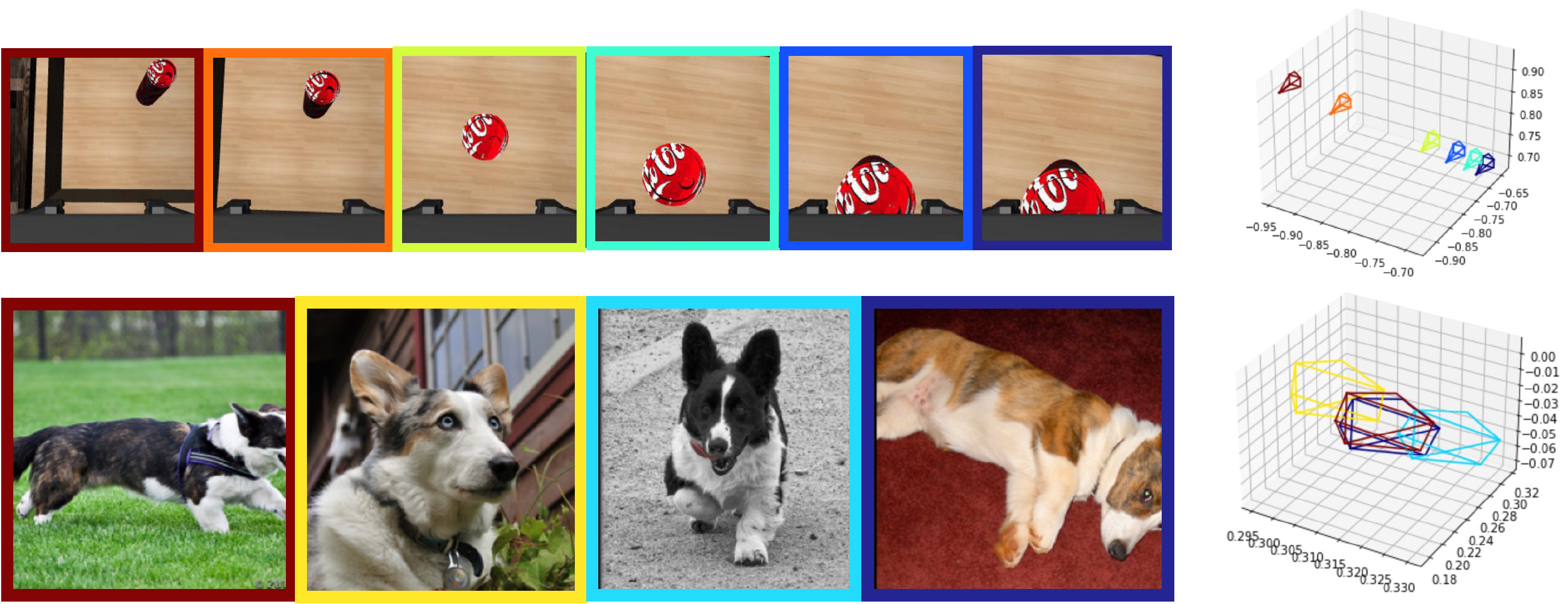}
    \caption{Visualization of image samples (left, with colored boundaries) and estimated camera positions in a 3D space (right). The color of the boundary on each image corresponds to the estimated camera from that image. \textbf{Top}: Images are from a video clip captured by an egocentric (eye-in-hand) camera on a robot arm in Can environment, ordered by timestep from left to right. The estimated camera positions approximately reflects the motion of the robot arm, which is moving towards right and down. \textbf{Bottom}: Samples from ImageNet-1K. The estimated camera pose of the second image (yellow boundary) is somehow at a head-up view, and the rest are at a top-down view. These estimated cameras are approximately aligned with human perception.}
    \label{fig:vis_cam}
\end{figure}

\begin{figure}[htbp]
    \centering
    \includegraphics[width=\textwidth]{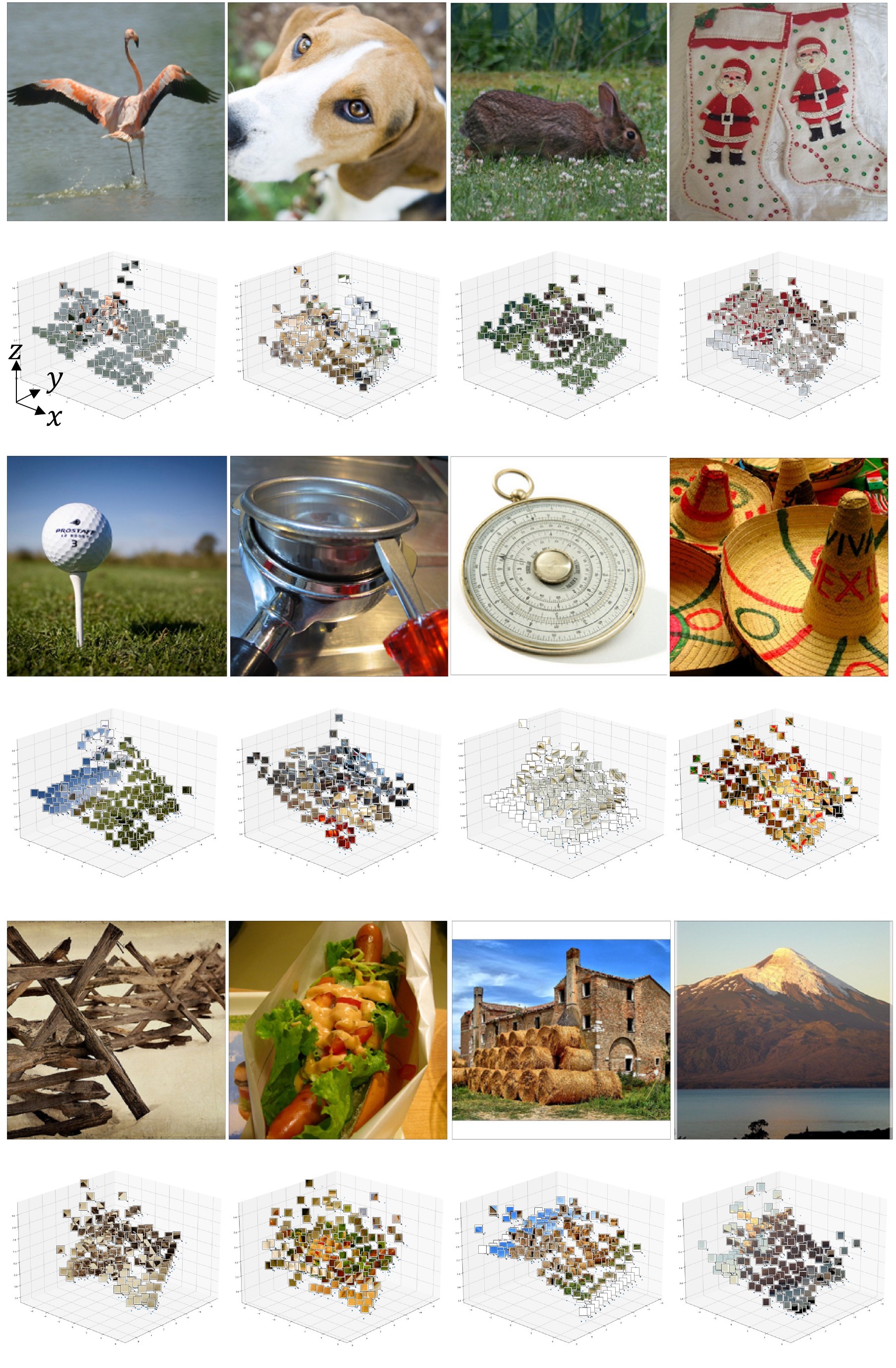}
    \caption{Visualization of image patches at their estimated 3D world locations. The center of the $xy$-plane (horizontal plane, at the bottom) is the origin of the 3D space. The vertical axis is $z$-axis. The patches corresponds to object-of-interest usually have larger $z$ values (corresponding to larger pseudo-depth values), which are localized ``farther'' from the origin. Most of the background patches have smaller $z$ values that are at the ``closer'' to the $xy$-plane.}
    \label{fig:vis_patch_3d}
\end{figure}

We also present more qualitative visualizations of estimated camera positions (Figure~\ref{fig:vis_cam}) and estimated 3D world locations of image patches (Figure~\ref{fig:vis_patch_3d}). We find that the estimations approximately reflect the ground truth or human perception which the model has no access to during training. These estimations are not necessarily required to be perfectly aligned with ground truth, but the results show that they are reasonable and sufficient for providing 3D information.

\section{Experiment on Perspective Augmentation}\label{sec:perspective_aug}
In this experiment, we investigate whether perspective augmentation applied to input images will help the model to learn viewpoint-agnostic representations. We test with DeiT+\modelname~on multi-view video alignment task. The training procedure is the same for all variants. Results are shown in Table~\ref{tab:perspective_aug}.
From the results we show that naively adding perspective augmentation does not improve the viewpoint-agnostic representation learning. Instead, it harms the performance compared to the DeiT baseline, since the perspective augmentation is overly artificial compared to the real-world viewpoint changes. Such augmentation does not contribute to viewpoint-agnostic representation learning.
\begin{table}[htbp]
    \centering
    \caption{Comparison between \modelname~and perspective augmentation on training data. Overall, perspective augmentation shows a negative effect on all the tasks, because the perspective augmentation on image is not the real viewpoint change.}
    \begin{tabular}{lrrrrr}
        \toprule
        Model &  Pouring & Pick & MC & Can & Lift\\
        \midrule
        DeiT & 0.426 & 0.244 & -0.115 & 0.789 & 0.716 \\
        DeiT + Perspective Augmentation & \textcolor{red}{0.200} & \textcolor{red}{-0.249} & \textcolor{red}{-0.419} & \textcolor{red}{0.342} & \textcolor{red}{0.486}\\
        DeiT + \modelname & \textbf{0.740} & \textbf{0.635} & \textbf{0.392} & \textbf{0.824} & \textbf{0.739} \\
        \bottomrule
    \end{tabular}
    \label{tab:perspective_aug}
\end{table}

\section{Discussion on Pseudo-depth Estimation}
\begin{figure}[htbp]
    \centering
    \includegraphics[width=0.7\textwidth]{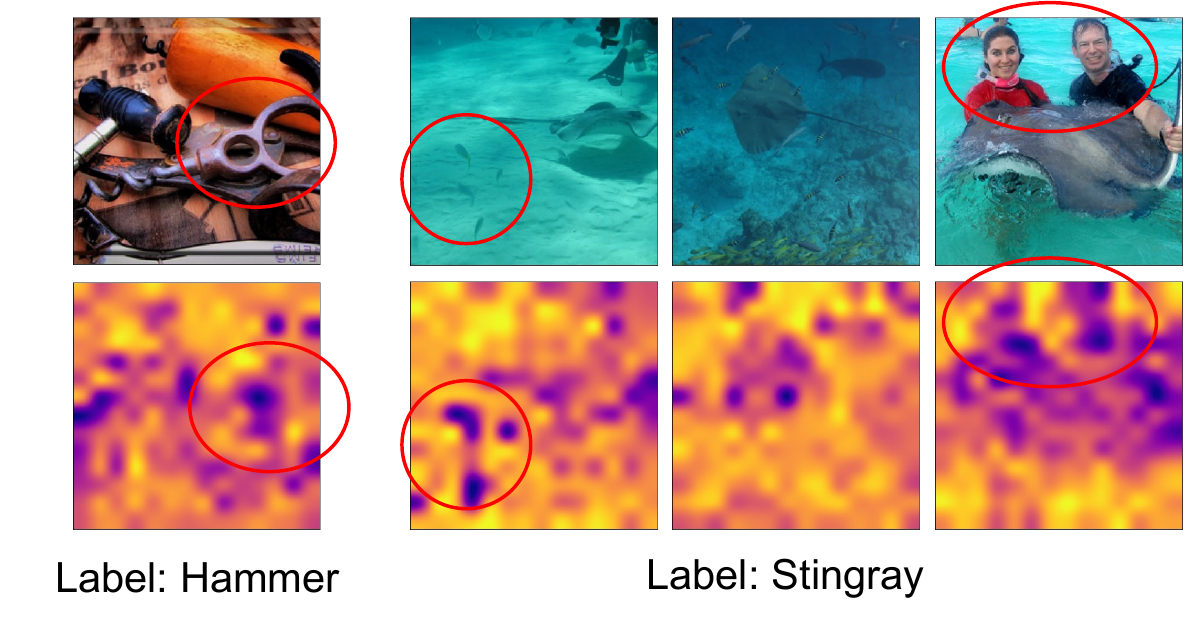}
    \caption{Examples of pseudo-depth estimation, applied to images with multiple types of objects. We use red circles to highlight the locations of non-class objects. In the hammer input, depth is also estimated on the other tool. In the first sample of stingray, three other fishes in the ``foreground'' corresponds to a lower depth value. In the last sample of stingray, two humans are predicted with a relative lower depth value, in between the stingray and the background.}
    \label{fig:depth_other_objects}
\end{figure}

Most of images from ImageNet have a simple scene (background), so it's easier for the pseudo-depth estimation to focus on objects. In examples shown in Figure~\ref{fig:depth_other_objects}, we show that the pseudo-depth is also estimated for other foreground objects apart from the primary class object.

\section{More Ablation Studies}
\subsection{How many \modelname s should be used?}
We explore the possibility of using multiple \modelname s in a Transformer backbone. This usage is essentially estimating 3D information from different levels of features, and injecting such information back into the backbone. We test several DeiT-T variants using multiple \modelname s on CIFAR-10 and results are shown in Table~\ref{tab:multiple_blocks}. We find that using multiple \modelname s further increases the performance in general compared to using only one \modelname. This shows the extra capacity from multiple \modelname s benefits representation learning. Specifically, we demonstrate that inserting \modelname s at layer 4, 6, and 8 yields the best result among all the strategies we explore. This experiment empirically shows multiple \modelname s potentially benefit the model.
\begin{table}[htbp]
    \centering
    \captionof{table}{CIFAR-10 Performance when using multiple \modelname s based on DeiT-T.}
    \begin{tabular}{lccccc}
    \toprule
        \textbf{\modelname~Location(s)} & N/A (DeiT baseline) & 4 & 4, 6, 8 & 4, 4, 4 & 2, 4, 6, 8 \\
        \textbf{CIFAR-10} & 74.1 & 78.8 & \textbf{79.5} & 79.3 & 79.1  \\
    \bottomrule
    \end{tabular}
    \label{tab:multiple_blocks}
\end{table}
\subsection{Regularization effect of \modelname}
Mixup~\cite{zhang2017mixup} and CutMix~\cite{yun2019cutmix} are commonly used image augmentation methods in training image models, which mixes two images to create an augmented image for training input. Such technique along with other augmentations provides diverse image samples so that the image model is regularized and avoids overfitting. We hypothesise that Mixup \& CutMix could potentially damage structures in original image, which may cause inaccurate depth estimation in \modelname, thus hamper the training procedure. Therefore, we conduct an experiment on ImageNet-1K of disabling Mixup \& CutMix. We train baseline DeiT and DeiT+\modelname~from scratch, and compare the results in both original validation set and perturbed set. In Table~\ref{tab:mixup_disable}, we find that both baseline and our method increases validation scores after disabling Mixup \& CutMix, and \modelname~still outperforms the baseline by 0.2\%. However, when tested on view-perturbed set, the baseline model shows a great performance drop (-6.8\%) which is much larger than \modelname~(-1.6\%). 
\begin{table}[htbp]
    \centering
    \caption{ImageNet-1K and ImageNet-1K-Perturbed results when Mixup \& Cutmix are disabled.}
    \begin{tabular}{lcc}
    \toprule
      \textbf{Model}  & \textbf{ImageNet-1K} & \textbf{ImageNet-1K-Perturbed}  \\
      \midrule
        DeiT-T & 73.4 & 61.3 \\
        DeiT-T + Mixup \& CutMix Disabled & 74.2 & \textcolor{debianred}{54.5} \\
        \cmidrule{1-3}
        DeiT-T+\modelname & 73.6 & 64.6 \\
        DeiT-T+\modelname~+ Mixup \& CutMix Disabled & 74.4 & \textcolor{applegreen}{63.0} \\
    \bottomrule
    \end{tabular}
    \label{tab:mixup_disable}
\end{table}
\subsection{Camera parameter estimation in video model}
We implemented and evaluated two different strategies for the camera parameter estimation $g(\cdot)$ in \modelname~for videos. These two strategies are: (a) Divided-Temporal (DT) and (b) Joint-Temporal (JT) estimation introduced below.
In (a) DT strategy, we estimate one set of camera parameters $[~\mathbf{R}|\mathbf{t}~]_t$ per input frame ($S_t$) in a dissociated manner, and thus estimate a total of $T$ camera matrices for the entire video.
In (b) JT strategy, we estimate only one camera from all frames $S=\{S_1, \dots, S_T\}$. The camera is shared across all spatial-temporal tokens associated with the latent 3D space. The underlying hypothesis is that the camera pose and location do not change during the video clip. JT could be helpful to properly constrain the model for scenarios where camera movement is not required, but it is not generalizable to scenarios where the subject of interest quite often moves within the field-of-view. 

By default, we used DT strategy in all experiments presented in the main paper. In Table~\ref{tab:dtjt}, we show a comparison of DT and JT strategies in different scenarios. 
Note that \modelname~implemented with JT strategy under-performs the baseline TimeSformer on Smarthome (in most of the CV2 experiments) dataset.
We find that the JT strategy adoption for video models is particularly effective when there is a large availability of training data, for example on NTU dataset. However, these results with JT strategy are inconsistent across different datasets and also less substantial w.r.t. the results with our default DT strategy. This shows the requirement of estimating camera matrix per frame rather than a global camera matrix for video representation tasks.

\begin{table}[htbp]
    \centering
    \caption{Comparison of DT and JT strategies in \modelname~ for action recognition task.}
    \resizebox{\textwidth}{!}{
    \begin{tabular}{lllllll}
        \toprule
        \multirow{2}{*}{\textbf{Method}} & Strategy & \multicolumn{2}{c}{\textbf{Smarthome} (CV2)}  & \multicolumn{2}{c}{\textbf{Smarthome} (CS)} & \textbf{NTU} (CV) \\ 
        \cmidrule(lr){3-4} \cmidrule(lr){5-6} \cmidrule(lr){7-7}
         & &  \textit{Acc} & \textit{mPA} & \textit{Acc} & \textit{mPA} & \textit{Acc} \\
         \midrule
         TimeSformer~\cite{timesformer} & - & 59.4 & 27.5 & 75.7 & 56.1 & 86.4 \\
         + \textbf{\modelname }& DT &\textbf{62.9}\perfinc{3.5} & \textbf{34.0}\perfinc{6.5} & 76.1\perfinc{0.4}  & 57.0\perfinc{0.9} & \textbf{87.9}\perfinc{1.5} \\
        + \modelname &   JT & 58.6\perfdec{0.8} & 30.9\perfinc{3.4} & \textbf{76.2}\perfinc{0.5} & \textbf{57.2}\perfinc{1.1} & \textbf{87.9}\perfinc{1.5} \\
         \midrule
         \multicolumn{7}{@{}l}{\makecell{\hspace{1.3in}\underline{\textbf{ Kinetics-400 pre-trained}}}} \vspace{0.1in}\\ 
         TimeSformer~\cite{timesformer} & - & 69.3 & 37.5 & 77.2 & 57.7 & 87.7 \\
         + \modelname~w/o K400  & DT &  69.5\perfinc{0.2} & 39.2\perfinc{1.7} & 77.5\perfinc{0.3} & 58.9\perfinc{1.2} & \textbf{88.8}\perfinc{1.1} \\
          \textbf{+ \modelname~w/ K400} & DT & \textbf{71.9}\perfinc{2.6} & \textbf{41.7}\perfinc{4.2} & \textbf{77.8}\perfinc{0.6} & \textbf{61.0}\perfinc{2.3} & 88.6\perfinc{0.9}\\
          + \modelname~w/o K400  & JT &  66.6\perfdec{2.7} & 35.0\perfdec{2.5} & 77.0\perfdec{0.2} & 58.6\perfinc{0.9} & 88.6\perfinc{0.9} \\
          + \modelname~w/ K400 & JT & 68.2\perfdec{0.9} & 37.1\perfdec{0.4} & 77.0\perfdec{0.2} & 59.9\perfinc{2.2} & 87.7\perfinc{0.0} \\
         \bottomrule
    \end{tabular}
    }
    \label{tab:dtjt}
\end{table}

\section{Limitations}
We find that \modelname~suffers when estimating small objects in the scene, or estimating objects in a complex scene, due to the coarse scale (in 16x16 image patches) from the backbone Transformer. One possible solution is to decrease the patch size or enlarge the input size in the backbone Transformer, but in practice it is computationally infeasible as Attention complexity grows quadratically. Similar problem occurs when \modelname~is applied on Transformers having hierarchical architectures like Swin Transformer~\cite{liu2021Swin}, where we find our improvement is minor compared to DeiT. In hierarchical architectures, image patches are merged after one stage so the resolution of the pseudo-depth map decreases quadratically. To solve this issue, we recommend to place \modelname~at the location before any patch merging in such hierarchical Transformers.

\section{Implementation Details}
\modelname~is easily inserted in Transformers.
The components of \modelname~are implemented by several MLPs and required geometric transformations in between. 
We keep the hidden dimension size in MLPs the same as the embedding dimensionality of Transformer backbone, Tiny=192, Small=384, Base=768 in specific.
We provide PyTorch-style pseudo-code about inserting \modelname~in Transformer (Algorithm~\ref{algo:3dtrl_in_transformer}) and about details of \modelname~(Algorithm~\ref{algo:3dtrl}). We use image Transformer for example and omit operations on \textit{CLS} token for simplicity. Full implementation including video model is provided in supplementary files.
\begin{algorithm}[htbp]
\SetAlgoLined
    \PyComment{Use \modelname~with Transformer backbone}\\
    \PyCode{Class Transformer\_with\_\modelname:}\\
    \Indp
        \PyCode{def \_\_init\_\_(self, config):}\\
        \Indp
            \PyComment{Initialize a Transformer backbone and \modelname}\\
            \PyCode{self.backbone = Transformer(config)} \\
            \PyCode{self.3dtrl = \modelname(config)}\\
            \PyComment{Before which Transformer layer we insert \modelname}\\
            \PyCode{self.3dtrl\_location = config.3dtrl\_location}\\
        \Indm
        ~\\
        \PyCode{def forward(self, tokens):}\\
        \Indp
            \PyCode{for i, block in enumerate(self.backbone.blocks):}\\
            \Indp
                \PyComment{Tokens go through \modelname~at desired insert location}\\
                \PyCode{if i == self.3dtrl\_location:}\\
                \Indp
                    \PyCode{tokens = self.3dtrl(tokens)}\\
                \Indm
                \PyComment{Tokens go through backbone layers}\\
                tokens = block(tokens)\\
            \Indm
            return tokens\\
        \Indm
    \Indm
\caption{PyTorch-style pseudo-code for using \modelname~in Transformer}
\label{algo:3dtrl_in_transformer}
\end{algorithm}
\begin{algorithm}[htbp]
\SetAlgoLined
    \PyCode{Class \modelname:}\\
    \Indp
        \PyComment{Make a \modelname}\\
        \PyCode{def \_\_init\_\_(self, config):}\\
        \Indp
            \PyComment{2D coordinates on image plane}\\
            \PyCode{self.u, self.v = make\_2d\_coordinates()}\\
            \PyComment{Depth estimator}\\
            \PyCode{self.depth\_estimator = nn.Sequential(}\\
            \Indp
                \PyCode{nn.Linear(config.embed\_dim, config.embed\_dim),}\\
                \PyCode{nn.ReLU(),}\\
                \PyCode{nn.Linear(config.embed\_dim, 1))}\\
            \Indm
            ~\\
            \PyComment{Camera parameter estimator, including a stem and two heads}\\
            \PyCode{self.camera\_estimator\_stem = nn.Sequential(}\\
            \Indp
                \PyCode{nn.Linear(config.embed\_dim, config.embed\_dim),}\\
                \PyCode{nn.ReLU(),}\\
                \PyCode{nn.Linear(config.embed\_dim, config.embed\_dim),}\\
                \PyCode{nn.ReLU(),}\\
                \PyCode{nn.Linear(config.embed\_dim, 32),}\\
                \PyCode{nn.ReLU(),}\\
                \PyCode{nn.Linear(32, 32))}\\
            \Indm
            \PyComment{Heads for rotation and translation matrices.}\\
            \PyCode{self.rotation\_head = nn.Linear(32, 3)}\\
            \PyCode{self.translation\_head = nn.Linear(32, 3)}\\
            ~\\
            \PyComment{3D positional embedding layer}\\
            \PyCode{self.3d\_pos\_embedding = nn.Sequential(}\\
            \Indp
                \PyCode{nn.Linear(3, config.embed\_dim),}\\
                \PyCode{nn.ReLU(),}\\
                \PyCode{nn.Linear(config.embed\_dim, config.embed\_dim))}\\
            \Indm
        \Indm
        ~\\
        \PyCode{def forward(self, tokens):}\\
        \Indp
            \PyComment{Depth estimation}\\
            \PyCode{depth = self.depth\_estimator(tokens)}\\
            \PyCode{camera\_centered\_coords = uvd\_to\_xyz(self.u, self.v, depth)}\\
            ~\\
            \PyComment{Camera estimation}\\
            \PyCode{interm\_rep = self.camera\_estimator\_stem(tokens)}\\
            \PyCode{rot, trans = self.rotation\_head(interm\_rep), self.translation\_head(interm\_rep)}\\
            \PyCode{rot = make\_rotation\_matrix(rot)}\\
            ~\\
            \PyComment{Transformation from camera-centered to world space}\\
            \PyCode{world\_coords = transform(camera\_centered\_coords, rot, trans)}\\
            ~\\
            \PyComment{Convert world coordinates to 3D positional embeddings}\\
            \PyCode{3d\_pos\_embed = self.3d\_pos\_embedding(world\_coords)}\\
            ~\\
            \PyComment{Generate output tokens}\\
            \PyCode{return tokens + 3d\_pos\_embed}
            
    \Indm
    
\caption{PyTorch-style pseudo-code for \modelname}
\label{algo:3dtrl}
\end{algorithm}

\section{Settings for Image Classification}
\paragraph{Datasets} For the task of image classification, we provide a thorough evaluation on three popular image datasets: CIFAR-10~\cite{cifar}, CIFAR-100~\cite{cifar}, and ImageNet~\cite{imagenet}. CIFAR-10/100 consists of 50k training and 10k test images, and ImageNet has 1.3M training and 50k validation images.
\paragraph{Training Configurations}
We follow the configurations introduced in DeiT~\cite{deit}. We provide a copy of configurations here in Table~\ref{tab:cifar_settings} (CIFAR) and Table~\ref{tab:in1k_settings} (ImageNet-1K) for reference. We use 4 NVIDIA Tesla V100s to train models with Tiny, Small and Base backbones on ImageNet-1K for $\sim$22 hours, $\sim$3 days and $\sim$5 days respectively.
\begin{table}[htbp]
    \centering
    \caption{CIFAR Training Settings}
    \begin{tabular}{lc}
    \toprule
        Input Size & 32$\times$32 \\
        Patch Size & 2$\times$2 \\
        Batch Size & 128 \\
        \midrule
        Optimizer & AdamW \\
        Optimizer Epsilon & 1.0e-06 \\
        Momentum & $\beta_1$, $\beta_2$ = 0.9, 0.999 \\
        layer-wise lr decay & 0.75 \\
        Weight Decay & 0.05 \\ 
        Gradient Clip & None \\
        \midrule
        Learning Rate Schedule & Cosine\\
        Learning Rate & 1e-3 \\
        Warmup LR & 1.0e-6 \\
        Min LR & 1e-6 \\
        Epochs & 50 \\
        Warmup Epochs & 5 \\
        Decay Rate & 0.988 \\
        drop path & 0.1 \\
        \midrule
        Exponential Moving Average (EMA) & True \\
        EMA Decay & 0.9999 \\
        \midrule
        Random Resize \& Crop Scale and Ratio & (0.08, 1.0), (0.67, 1.5)\\
        Random Flip & Horizontal 0.5; Vertical 0.0 \\
        Color Jittering & None \\
        Auto-agumentation & rand-m15-n2-mstd1.0-inc1\\
        Mixup & True \\
        Cutmix & True \\
        Mixup, Cutmix Probability & 0.8, 1.0 \\
        Mixup Mode & Batch \\
        Label Smoothing & 0.1 \\
    \bottomrule
    \end{tabular}
    \label{tab:cifar_settings}
\end{table}
\begin{table}[htbp]
    \centering
    \caption{ImageNet-1K Training Settings~\cite{deit}}
    \begin{tabular}{lc}
    \toprule
        Input Size & 224$\times$224 \\
        Crop Ratio & 0.9  \\
        Batch Size & 512 \\
        \midrule
        Optimizer & AdamW \\
        Optimizer Epsilon & 1.0e-06 \\
        Momentum & 0.9 \\
        Weight Decay & 0.3 \\ 
        Gradient Clip & 1.0 \\
        \midrule
        Learning Rate Schedule & Cosine\\
        Learning Rate & 1.5e-3 \\
        Warmup LR & 1.0e-6 \\
        Min LR & 1.0e-5 \\
        Epochs & 300 \\
        Decay Epochs & 1.0 \\
        Warmup Epochs & 15 \\
        Cooldown Epochs & 10 \\
        Patience Epochs & 10 \\
        Decay Rate & 0.988 \\
        \midrule
        Exponential Moving Average (EMA) & True \\
        EMA Decay & 0.99992 \\
        \midrule
        Random Resize \& Crop Scale and Ratio & (0.08, 1.0), (0.67, 1.5)\\
        Random Flip & Horizontal 0.5; Vertical 0.0 \\
        Color Jittering & 0.4 \\
        Auto-agumentation & rand-m15-n2-mstd1.0-inc1\\
        Mixup & True \\
        Cutmix & True \\
        Mixup, Cutmix Probability & 0.5, 0.5 \\
        Mixup Mode & Batch \\
        Label Smoothing & 0.1 \\
    \bottomrule
    \end{tabular}
    \label{tab:in1k_settings}
    \vspace{-10pt}
\end{table}

\section{Settings for Video Alignment}
\paragraph{Datasets}
We provide the statistics of 5 datasets used for video alignment in Table~\ref{tab:video_align_statistics}. In general, datasets with fewer training videos, more/diverse viewpoints, and longer videos are harder for alignment. We will also provide the copy of used/converted dataset upon publish.
\begin{table}[htbp]
    \centering
    \caption{Statistics of multi-view datasets used for video alignment. }
    \vspace{3pt}
    \begin{tabular}{lccc}
    \toprule
        \textbf{Dataset} & \textbf{\# Training/Validation/Test Videos} & \textbf{\# Viewpoints} & \textbf{Average Frames/Video}  \\
        \midrule
        Pouring & 45 / 10 / 14 & 2 & 266\\
        MC & 4 / 2 / 2 & 9 & 66\\
        Pick & 10 / 5 / 5 & 10 & 60\\
        Can & 200 / 50 / 50  & 5 & 38\\
        Lift & 200 / 50 / 50  & 5 & 20\\
    \bottomrule
    \end{tabular}
    \label{tab:video_align_statistics}
    \vspace{-10pt}
\end{table}
\paragraph{Training Configurations}
The training setting for video alignment is listed in Table~\ref{tab:video_align_setting}. The setting is the same for all datasets and all methods for fair comparison. GPU hours required for training vary across datasets, depending on the size of datasets and early stopping (convergence). Approximately we use 24 hours in total to fully train on all 5 datasets using an NVIDIA RTX A5000.
\begin{table}[htbp]
    \centering
    \caption{Training Settings for Video Alignment}
    \begin{tabular}{lc}
    \toprule
     Positive Window of TCN Loss & 3 frames in MC, Pick, Pouring; 2 frames in Can and Lift\\
     Learning Rate    & 1e-6 \\
     Batch Size     & 1 \\
     Optimizer & Adam \\
     Gradient Clip & 10.0 \\
     Early Stopping & 10 epochs \\
     Random Seed & 42 \\
     Augmentations & No \\
    \bottomrule
    \end{tabular}
    \label{tab:video_align_setting}
    \vspace{-10pt}
\end{table}
\section{Settings for Video Representation Learning}\label{sec:video_exp_setting}
\paragraph{Datasets} Our dataset choices are based on multi-camera setups in order to provide cross-view evaluation. Therefore, we evaluate the effectiveness of \modelname on two multi-view datasets Toyota Smarthome~\cite{smarthome} and NTU-RGB+D~\cite{NTU_RGB+D}. We also use Kinetics-400~\cite{kinetics} for pre-training the video backbone before plugging-in \modelname.

\noindent Toyota-Smarthome (Smarthome) is a recent ADL dataset recorded in an apartment where 18 older subjects carry out tasks of daily living during a day. The dataset contains 16.1k video clips, 7 different camera views and 31 complex activities performed in a natural way without strong prior instructions. For evaluation on this dataset, we follow cross-subject ($CS$) and cross-view ($CV_2$) protocols proposed in~\cite{smarthome}. We ignore protocol $CV_1$ due to limited training samples. 

\noindent NTU RGB+D (NTU) is acquired with a Kinect v2 camera and consists of 56880 video samples with 60 activity classes. The activities were performed by 40 subjects and recorded from 80 viewpoints. For each frame, the dataset provides RGB, depth and a 25-joint skeleton of each subject in the frame. For evaluation, we follow the two protocols proposed in~\cite{NTU_RGB+D}: cross-subject (CS) and cross-view (CV).

\noindent Kinetics-400 (K400) is a large-scale dataset with ~240k training, 20k validation and 35k testing videos in 400 human action categories. However, this dataset do not posses the viewpoint challenges, we are addressing in this paper. So, we use this dataset only for pre-training purpose as used by earlier studies.

\paragraph{Training Configurations}
We use clips of size $8 \times 224 \times 224 \times 3$, with frames sampled at a rate of 1/32. 
We use a ViT-B encoder with patch size $16 \times 16$. 
The training setting for action recognition on both datasets follow the configurations provided in~\cite{timesformer}. We train all the video models on 4 RTX 8000 GPUs with a batch size of 4 per GPU for 15 epochs. A gradient accumulation is performed to have an effective batch size of 64. Similar to~\cite{timesformer}, we train our video models with SGD optimiser with $0.9$ momentum and $1e-4$ weight decay.
During inference, we sample one and 10 temporal clips from the entire video on NTU and Smarthome datasets respectively. We use 3 spatial crops (top-left, center, bottom-right) from each temporal clip and obtain the final prediction by averaging the scores for all the crops. 
\section{More Pseudo-depth Estimation Visualization}
\begin{figure}[htbp]
    \centering
    \includegraphics[width=0.89\textwidth]{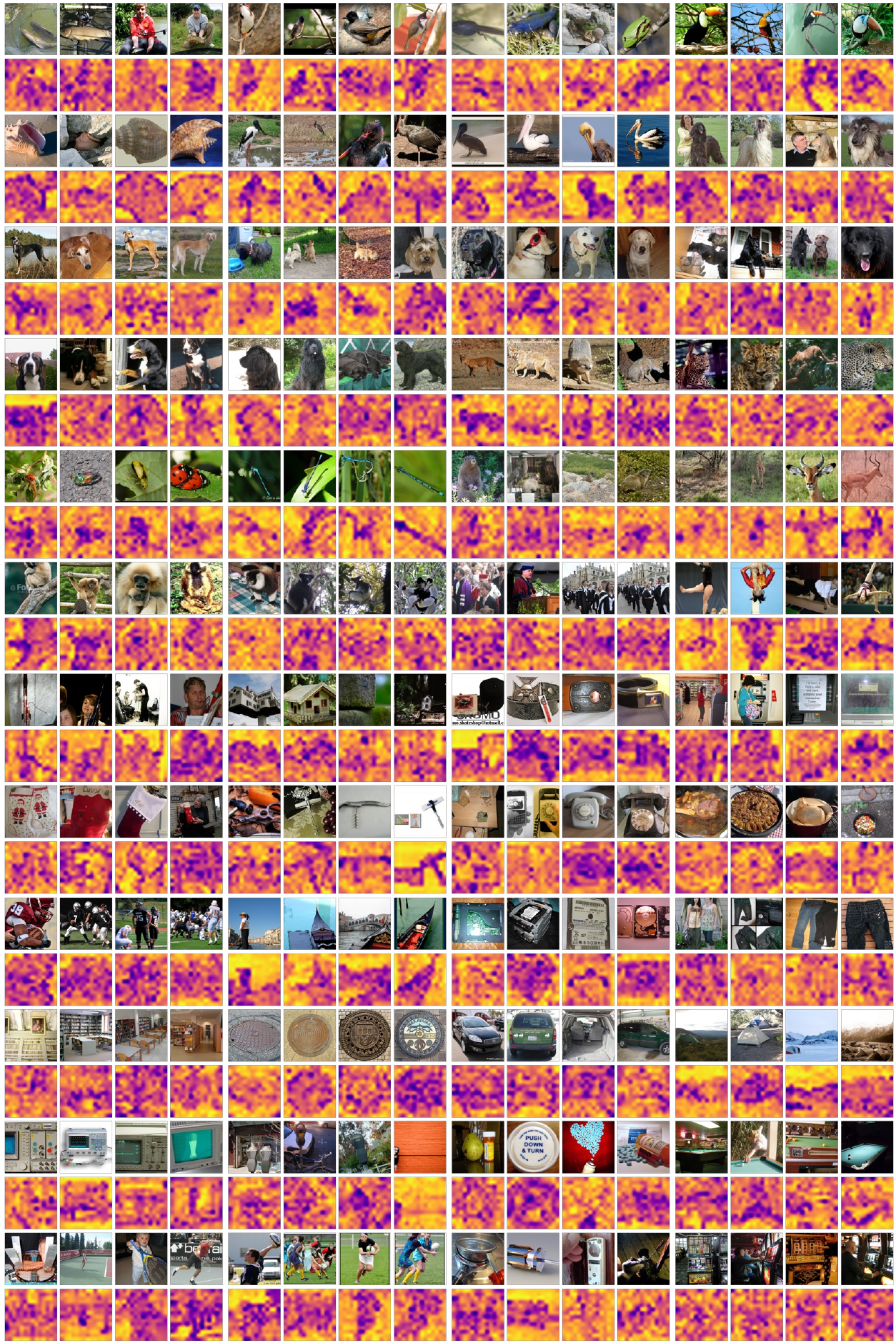}
    \caption{More examples of pseudo-depth maps.}
    \label{fig:morevisdepth}
\end{figure}

Figure~\ref{fig:morevisdepth} gives examples of more pseudo-depth maps.
\newpage
\section{Pseudo-depth Map Visualization over Training Epochs}
\begin{figure}[htbp]
    \centering
    \includegraphics[width=\textwidth]{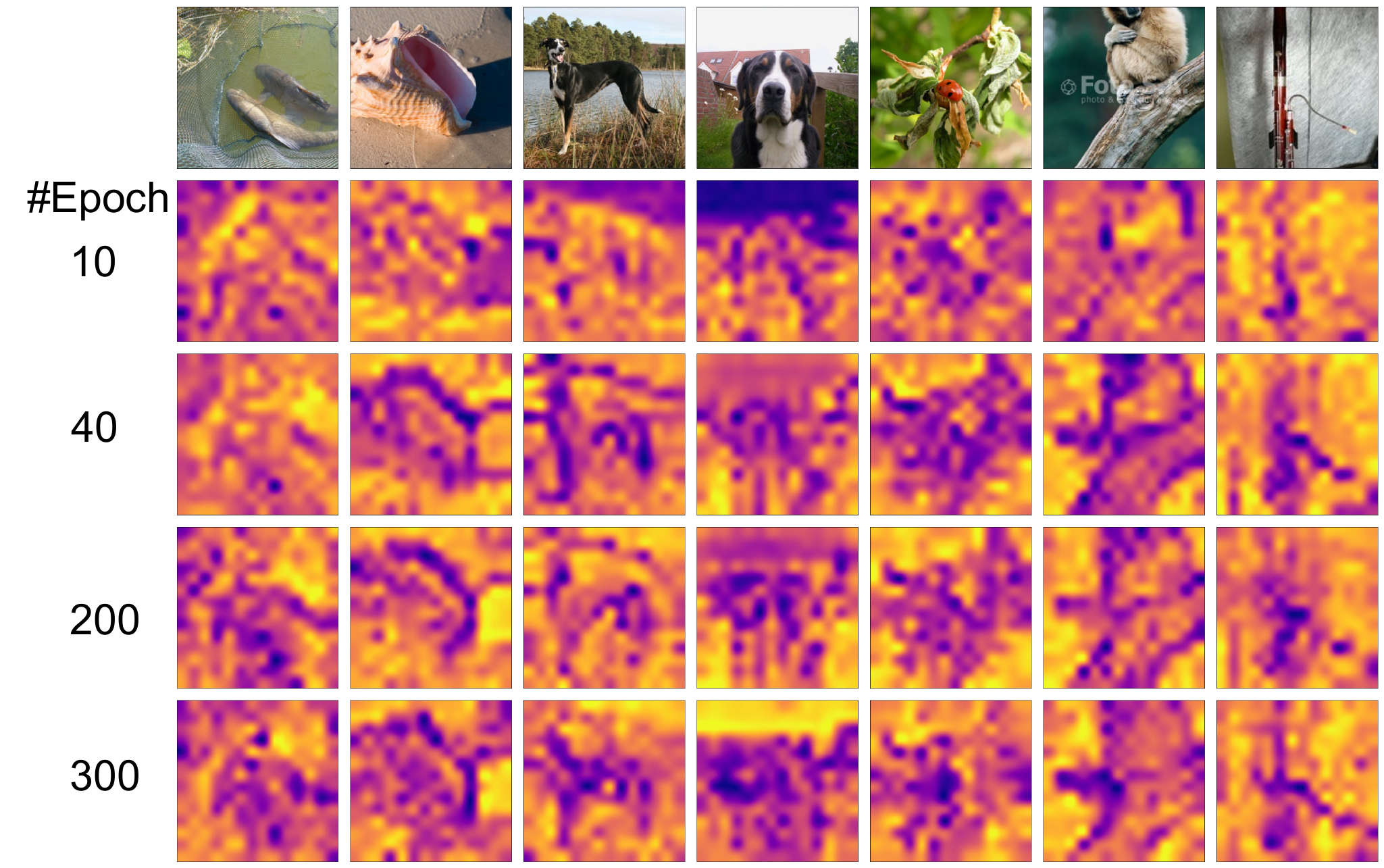}
    \caption{Pseudo-depth maps over training epochs.}
    \label{fig:visdepthepoch}
\end{figure}

Figure~\ref{fig:visdepthepoch} gives examples of pseudo-depth estimation over Training Epochs. We note that the results are from training \modelname~with IN-1K. We find that the estimation varies significantly from epoch 10 to epoch 40 (higher foreground-background correctness, less missing parts of objects), but changes only a bit from epoch 40 to epoch 200 and finally to epoch 300 (mostly scales). This observation is also coherent with our quantitative evaluation in Section~\ref{sec:depth_eval}. Thus, the pseudo-depth estimation learns promptly, however the model convergence takes longer time since we are optimizing for a downstream task (eg. classification).


\end{document}